\documentclass[letterpaper,twocolumn,10pt]{article}
\usepackage{mystyle}

\usepackage{booktabs, siunitx, multirow, multicol, tabularx}
\usepackage{amsmath,amsthm,amsfonts,amssymb,mathtools}
\usepackage{enumerate,enumitem}
\usepackage{graphicx,graphics}
\usepackage{lipsum}
\usepackage{tikz}
\usepackage{float}
\usepackage{comment}
\usepackage[utf8]{inputenc}
\DeclarePairedDelimiter{\ceil}{\lceil}{\rceil}
\usepackage{xspace}
\usepackage[noend, ruled]{algorithm2e}

\usepackage{caption,subcaption}
\usepackage{algpseudocode}
\usepackage{colortbl}
\definecolor{mygray}{gray}{.9}
\usepackage[absolute]{textpos}

\definecolor{revision}{RGB}{0,0,0}

\newcommand{\mypara}[1]{\noindent\textbf{#1.}\xspace}

\renewcommand{\Pr}[1]{\ensuremath{\mathsf{Pr}\left[#1\right]}\xspace}
\newcommand{\tuple}[1]{\ensuremath{\langle #1 \rangle}}

\newcommand{\sysname}{\ensuremath{\mathsf{GraphEraser}}\xspace}

\newcommand{\optaggr}{\ensuremath{\mathsf{LBAggr}}\xspace}
\newcommand{\majaggr}{\ensuremath{\mathsf{MajAggr}}\xspace}
\newcommand{\meanaggr}{\ensuremath{\mathsf{MeanAggr}}\xspace}

\newcommand{\scratch}{\ensuremath{\mathsf{Scratch}}\xspace}
\newcommand{\sisa}{\ensuremath{\mathsf{SISA}}\xspace}
\newcommand{\random}{\ensuremath{\mathsf{Random}}\xspace}
\newcommand{\lpa}{\ensuremath{\mathsf{LPA}}\xspace}
\newcommand{\ekm}{\ensuremath{\mathsf{EKM}}\xspace}
\newcommand{\blpa}{\ensuremath{\mathsf{BLPA}}\xspace}
\newcommand{\bkm}{\ensuremath{\mathsf{BEKM}}\xspace}

\newcommand{\gin}{\ensuremath{\mathsf{GIN}}\xspace}
\newcommand{\sage}{\ensuremath{\mathsf{SAGE}}\xspace}
\newcommand{\gat}{\ensuremath{\mathsf{GAT}}\xspace}
\newcommand{\gcn}{\ensuremath{\mathsf{GCN}}\xspace}

\newcommand{\model}{\ensuremath{\mathcal{F}}\xspace}
\newcommand{\attack}{\ensuremath{\mathcal{A}}\xspace}
\newcommand{\dset}{\ensuremath{\mathcal{D}}\xspace}
\newcommand{\post}{\ensuremath{P}\xspace}

\newcommand{\graph}{\ensuremath{\mathcal{G}}\xspace}
\newcommand{\nodeset}{\ensuremath{\mathcal{V}}\xspace}
\newcommand{\edge}[1]{\ensuremath{e_{#1}}\xspace}
\newcommand{\adj}{\ensuremath{A}\xspace}
\newcommand{\embed}{\ensuremath{E}\xspace}
\newcommand{\embedset}{\ensuremath{\mathbb{\embed}}\xspace}
\newcommand{\feat}{\ensuremath{X}\xspace}
\newcommand{\neigh}[1]{\ensuremath{\mathcal{N}_{#1}}\xspace}

\newcommand{\aggr}{\ensuremath{\Phi}\xspace}
\newcommand{\upd}{\ensuremath{\Psi}\xspace}
\newcommand{\aggregate}[2]{\ensuremath{\aggr^{#1}\left( #2 \right)}\xspace}
\newcommand{\update}[2]{\ensuremath{\upd^{#1}\left( #2 \right)}\xspace}
\newcommand{\messagee}{\ensuremath{\mathbf{m}}\xspace}

\newcommand{\loss}{\ensuremath{\mathcal{L}}\xspace}

\newcommand{\community}{\ensuremath{\mathbb{C}}\xspace}
\newcommand{\centroid}{\ensuremath{C}\xspace}
\newcommand{\neighcount}{\ensuremath{\xi}\xspace}
\newcommand{\comreassign}{\ensuremath{\mathbb{F}}\xspace}

\begin{document}

\begin{textblock}{16}(1.5,1)
To Appear in 2022 ACM SIGSAC Conference on Computer and Communications Security, November 7-11, 2022.
\end{textblock}

\title{\Large \bf Graph Unlearning}
\date{}

\author{
Min Chen\textsuperscript{1}\ \ \
Zhikun Zhang\textsuperscript{1}\thanks{Corresponding author.}\ \ \
Tianhao Wang\textsuperscript{2}\ \ \
Michael Backes\textsuperscript{1}\ \ \ 
\\
Mathias Humbert\textsuperscript{3}\ \ \
Yang Zhang\textsuperscript{1}
\\
\textsuperscript{1}\textit{CISPA Helmholtz Center for Information Security} \ \ \ 
\\
\textsuperscript{2}\textit{University of Virginia} \ \ \
\textsuperscript{3}\textit{University of Lausanne} \ \ \ 
}

\maketitle

\begin{abstract}
\textit{Machine unlearning} is a process of removing the impact of some training data from the machine learning (ML) models upon receiving removal requests.
While straightforward and legitimate, retraining the ML model from scratch incurs a high computational overhead.
To address this issue, a number of approximate algorithms have been proposed in the domain of image and text data, among which \sisa is the state-of-the-art solution.
It randomly partitions the training set into multiple shards and trains a constituent model for each shard.
However, directly applying \sisa to the graph data can severely damage the graph structural information, and thereby the resulting ML model utility.
In this paper, we propose \sysname, a novel machine unlearning framework tailored to graph data.
Its contributions include two novel graph partition algorithms and a learning-based aggregation method.
We conduct extensive experiments on five real-world graph datasets to illustrate the unlearning efficiency and model utility of \sysname.
It achieves 2.06$\times$ (small dataset) to 35.94$\times$ (large dataset) unlearning time improvement.
On the other hand, \sysname achieves up to $62.5\%$ higher F1 score and our proposed learning-based aggregation method achieves up to $112\%$ higher F1 score.\footnote{Our code is available at \url{https://github.com/MinChen00/Graph-Unlearning}.}
\end{abstract}

\section{Introduction}
\label{section:introduction}
Data protection has attracted increasing attentions recently, and several regulations have been proposed to protect the privacy of individual users, such as the General Data Protection Regulation (GDPR)~\cite{GDPR} in the European Union, the California Consumer Privacy Act (CCPA)~\cite{CCPA} in California, the Personal Information Protection and Electronic Documents Act (PIPEDA)~\cite{PIPEDA} in Canada, and the Brazilian General Data Protection Law (LGPD) in Brazil~\cite{LGPD}.
One of the most important and controversial articles in these regulations is \textit{the right to be forgotten}, which entitles data subject the right to delete their data from an entity storing it.
In the context of machine learning (ML), researchers have argued that, under the requirement of the right to be forgotten, the \textit{model provider} has the obligation to eliminate any impact of the data whose owner requested to be forgotten, which is referred to as \textit{machine unlearning}~\cite{CY15,BCCJTZLP21}.

A deterministic machine unlearning approach consists in removing the revoked sample and retraining the ML model from scratch.
Given that this method can be computationally prohibitive when the underlying dataset is large, many approximate machine unlearning methods have been proposed~\cite{CY15, GGVZ19,CYASMY18,GAS20,GGHM20,ISCZ21,BCCJTZLP21,NRS21}.
Among the existing solutions, the \sisa \xspace (\underline{S}harded, \underline{I}solated, \underline{S}liced, and \underline{A}ggregated) is the most general one in terms of model architecture~\cite{BCCJTZLP21}.
The basic idea of \sisa is to randomly split the training dataset into several disjoint shards and train each shard model separately.
Upon receiving an unlearning request, the model provider only needs to retrain the corresponding shard model.

\sisa has been designed to handle image and text data in the Euclidean space.
However, numerous important real-world datasets are represented in the form of graphs, such as social networks~\cite{QTMDWT18}, financial networks~\cite{WDCWBRL19a}, biological networks~\cite{GN02}, recommender systems~\cite{PEZZRL20,YHCEHL18}, and transportation networks~\cite{DWZLXH19}.
In order to take advantage of the rich information contained in graphs, a new family of ML models, graph neural networks (GNNs), has been recently proposed and has already shown great promise~\cite{KW17,VCCRLB18,TA19,PEZZRL20,DWZLXH19,HBJPAABCMS20,WDD20,BL21}.
The core idea of GNNs is to transform the graph data %
into low-dimensional vectors by aggregating the feature information from neighboring nodes.
Similar to other ML models, GNNs can be trained on sensitive graphs such as social networks~\cite{PEZZRL20,QTMDWT18,SHZCYBZS22}, where the data subject may request to revoke their data.
However, learning representative GNNs rely on graph structural information. Randomly partitioning the nodes into sub-graphs as in \sisa could severely damage the resulting model utility.
Therefore, there is a pressing need for novel methods for unlearning previously seen -- but revoked --  data samples in the context of GNNs.

\mypara{Our Contributions}
In this paper, we propose \sysname, an efficient unlearning framework to achieve high unlearning efficiency and reserve high model utility in GNNs.
Concretely, we first identify two common types of machine unlearning requests in the context of GNNs, namely \emph{node unlearning} and \emph{edge unlearning}.
We then propose a general pipeline for machine unlearning in GNN models.

To permit efficient retraining while keeping the structural information of the graph, we propose two graph partition strategies.
The first strategy focuses on the graph structural information and tries to preserve it to the greatest extent by relying on community detection.
Our second strategy takes both graph structural and node feature information into consideration.
In order to keep both pieces of information, we transform the node features and graph structure into embedding vectors that we then cluster into different shards.
However, a graph partitioned by traditional community detection\cite{ZN15,RB08,RAK07,WZ08} and clustering methods might lead to highly unbalanced shard sizes due to the structural properties of real-world graphs~\cite{ZN15, GN02}.
In such case, many (if not most) of the revoked samples would belong to the largest shard whose retraining time would be substantial and the unlearning process would then become highly inefficient.
We propose a general principle for balancing the shards resulting from the graph partition and instantiating it with two novel balanced graph partition algorithms to avoid this situation.
In addition, considering that the different shard models do not uniformly contribute to the final prediction, we further propose a learning-based aggregation method that optimizes the importance score of the shard models to eventually improve the global model utility.

We show in the experimental results that \sysname can effectively improve the unlearning efficiency.
For instance, the average unlearning time is up to $2.06\times$ shorter on the smallest dataset and up to $35.94\times$ shorter on the largest dataset compared to retraining from scratch, while \sysname achieves comparable model utility with retraining from scratch.
In addition, \sysname also provides an advanced model utility than random partitioning.
Concretely, \sysname achieves up to $62.5\%$ higher F1 score than that of random partitioning.
Furthermore, our learning-based aggregation method can effectively improve the model utility compared to the mean and majority-vote aggregation methods.
Our proposed learning-based aggregation achieves up to $93\%$ higher F1 score than that of the mean aggregation and $112\%$ higher F1 score than that of the majority vote aggregation .

In summary, we make the following contributions.
\begin{itemize}[leftmargin=*]
    \item To the best of our knowledge, \sysname is the first approach that addresses the machine unlearning problem for GNN models.
    Concretely, we formally define two types of machine unlearning requests in the context of GNN and propose a general pipeline for graph unlearning.
    \item We propose a unified principle to achieve balanced graph partitioning and instantiate it with two balanced graph partition algorithms.
    \item To improve the model utility resulting from \sysname, we propose a learning-based aggregation method.
    \item We conduct extensive experiments on five real-world graph datasets and four state-of-the-art GNN models to illustrate the unlearning efficiency and model utility resulting from \sysname.
\end{itemize} 

\mypara{Roadmap}
The rest of this paper is organized as follows.
We introduce the background knowledge of GNNs and machine unlearning in \autoref{section:background}.
In \autoref{section:graph_unlearning}, we formally define graph unlearning and introduce the general pipeline for \sysname.
\autoref{section:graph_partition} introduces the graph partition module for \sysname.
\autoref{section:optimal_aggregation} introduces the learning-based aggregation method.
We perform extensive experiments in \autoref{section:experiment} to evaluate the unlearning efficiency and model utility of \sysname.
We discuss some special cases in \autoref{section:discussion}.
We summarize the related work in \autoref{section:related} and conclude the paper in \autoref{section:conclusion}.

\section{Preliminaries}
\label{section:background}

\subsection{Graph Neural Networks}
\label{subsection:gnn}
We denote an attributed graph by $\graph = \tuple{\nodeset, \adj, \feat}$, where \nodeset is the set of all nodes, $\adj \in \{ 0, 1\}^{n \times n}$ is the adjacency matrix ($n = |\nodeset|$), and $\feat \in \mathbb{R}^{n \times d_\feat}$ is the feature matrix with $d_\feat$ being the dimension of node features.
We further denote \edge{u, v} as an edge between $u$ and $v$ in \graph.
We summarize the frequently used notations in \autoref{table:notations}. 

The high-level intuition behind GNNs is that the neighboring nodes in a graph tend to have similar features. The GNN models are designed so that each node can aggregate information from its neighbors and form an embedding (e.g., a size-512 vector).
The node embeddings can then be used to conduct downstream tasks, such as node classification~\cite{HYL17, VCCRLB18, PEZZRL20}, link prediction~\cite{WXCY17, ZC17}, and graph classification~\cite{XHLJ19, WHX19}.

In this paper, we focus on the \emph{node classification task}, whose goal is to use a GNN to predict the label of a node $u \in \nodeset$ given the node's features $X_u$ and its neighbors' information.
To train a GNN, we rely on the technique called \textit{message passing}, we refer the readers to \autoref{app:gnn_details} for more details.

\begin{table}[!t]
\centering
\caption{Summary of the notations.}
\resizebox{0.9\linewidth}{!}{ 
\vspace{-0.5cm}
\setlength{\tabcolsep}{0.9em}
{
\begin{tabular}{l|l}
\toprule 
\textbf{Notation} & \textbf{Description} \\ \midrule
	  $\graph = \tuple{\nodeset, \adj, \feat}$   & Graph   \\ 
      $u, v \in \nodeset$ & Nodes in \graph \\
      $\edge{u, v}$ & Edge that connects $u$ and $v$ \\
      \adj & Adjacency matrix of \graph \\
      \feat &  Attributes associated with \nodeset \\ 
      \neigh{u} & Neighborhood nodes of $u$ \\ 
	  $\embed_u$ & Node embedding of $u$ \\ 
	  $d_{\feat}$ / $d_{\embed}$ & Dimension of attributes / embeddings \\ 
      \aggr & Aggregation operation in message passing \\
      \upd & Updating operation in message passing\\
      \messagee & Message received from neighbors \\
\bottomrule	 
\end{tabular}
}
}
\label{table:notations}
\end{table}

\begin{figure*}[!t]
\begin{center}
\includegraphics[width=0.75\textwidth]{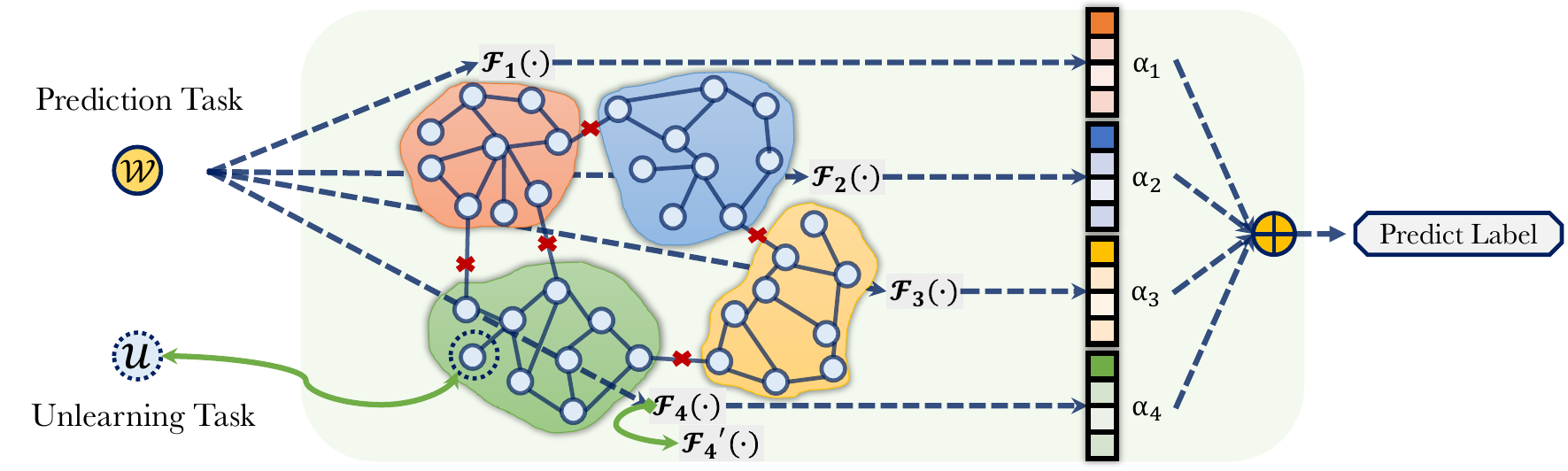}
\end{center}
\caption{A schematic view of the framework of \sysname. 
It partitions the original training graph into disjoint shards, parallelly trains a set of shard models $\model_i$, and learns an optimal importance score $\alpha_i$ for each shard model.
When a node $w$ needs prediction, \sysname sends $w$ to all the shard models and obtains the corresponding posteriors, which are then aggregated using the optimal importance score $\alpha_i$ to make the prediction.
When a node $u$ mounts an unlearning request, \sysname removes $u$ from the corresponding shard and retrains the shard model.
}
\label{figure:pipeline}
\end{figure*}

\subsection{Machine Unlearning}
Thanks to new legislation ensuring the ``right to be forgotten'', individuals can now formally request the deletion of their data from the service provider (or the data controller).
In the ML context, this implies that the model provider should delete the revoked sample from its training set. 
Still, it should also eliminate any influence of the revoked sample on the resulting ML model.

\mypara{Retraining from Scratch}
The most direct way to implement unlearning is to delete the revoked sample and retrain the ML model from scratch by using as training set the original dataset without the deleted sample.
Retraining from scratch is an effective and easy-to-enforce method for unlearning.
However, when the model is complex and the original training dataset is large, the computational overhead of retraining becomes prohibitive. 
To reduce the computational overhead, many approximation approaches have been proposed~\cite{ISCZ21,CY15,GAS20,BCCJTZLP21}, among which \sisa~\cite{BCCJTZLP21} is the most flexible one in terms model architecture.

\mypara{SISA}
SISA refers to \underline{S}harded, \underline{I}solated, \underline{S}liced, and \underline{A}ggregated, which is an ensemble learning-based method that can handle different ML model architectures.
With this approach, the training set $\dset_o$ is first partitioned into $k$ disjoint shards $\dset_o^1, \dset_o^2, \cdots, \dset_o^k$.
These $k$ shards are then used separately to train a set of ML models $\model_o^1, \model_o^2, \cdots, \model_o^k$.
At inference time, the $k$ individual predictions from the different shard models are simply aggregated (e.g., with majority voting) to provide a global prediction.
When the model owner receives a request to delete a new data sample, it just needs to retrain the shard model whose shard contains this sample, leading to a significant time gain with respect to retraining the whole model from scratch.

\section{Graph Unlearning}
\label{section:graph_unlearning}

\subsection{Problem Definition}
\label{subsection:problem_definition}
In the context of GNNs, the training set $\dset_o$ is in the form of a graph $\graph_o$, and a sample $x \in \dset_o$ corresponds to a node $u \in \graph_o$.
For presentation purposes, we use \textit{training graph} to represent \textit{training set} in the rest of this paper.
We identify two types of machine unlearning scenarios in the GNN setting, namely \emph{node unlearning} and \emph{edge unlearning}.

\mypara{Node Unlearning}
For a trained GNN model $\model_o$, the data of each data subject corresponds to a node in the GNN's training graph $\graph_o$.
In node unlearning, when a data subject $u$ asks the model provider to revoke all their data, this means the model provider should unlearn $u$'s node features and their links with other nodes from the GNN's training graph.
Taking social network as an example, node unlearning means a user's profile information and social connections need to be deleted from the training graph of a target GNN.
Formally, for node unlearning with respect to a node $u$, the model provider needs to obtain an unlearned model $\model_u$ trained on $\graph_u = \graph_o \setminus \{\feat_u, \edge{u, v} | \forall v \in \neigh{u}\}$, where $\feat_u$ is the feature vector of $u$.

\mypara{Edge Unlearning}
In edge unlearning, a data subject wants to revoke one edge between their node $u$ and another node $v$.
Still using social network as an example, edge unlearning means a social network user wants to hide their relationship with another individual.
Formally, to respond to the unlearning request for $\edge{u, v}$, the model provider needs to obtain an unlearned model $\model_u$ trained on $\graph_u = \graph_o \setminus \{ \edge{u, v} | v \in \neigh{u} \}$.
The features of the two nodes remain in the training graph.

\mypara{General Unlearning Objectives}
In the design of machine unlearning algorithms, we consider two major factors: \textit{unlearning efficiency} and \textit{model utility}.
The former is related to the retraining time when receiving an unlearning request.
This time should be as short as possible.
The latter is related to the unlearned model's prediction accuracy.
Ideally, the prediction accuracy should be close to retraining from scratch.
In summary, the unlearning algorithm should satisfy two general objectives: \textit{High Unlearning Efficiency} and \textit{Comparable Model Utility}.

\mypara{Challenges of Unlearning in GNNs}
As mentioned before, the state-of-the-art approach for machine unlearning is \sisa~\cite{BCCJTZLP21}, which randomly partitions the training set into multiple shards and trains a constituent model for each shard.
\sisa has shown to achieve high unlearning efficiency and comparable model utility for ML models whose inputs reside in the Euclidean space, such as images and texts.
However, the input of a GNN is a graph, and data samples, i.e., nodes of the graph, are not independent identically distributed.
Naively applying \sisa on GNNs for unlearning, i.e., randomly partitioning the training graph into multiple shards, will destroy the training graph's structure which may result in large model utility loss.
One solution is to rely on community detection methods to partition the training graph by the detected communities, which can preserve the graph structure to a large extent.
However, directly adopting classical community detection methods may lead to highly unbalanced shards in terms of shard size due to the specific structural properties of real-world graphs~\cite{GN02,RAK07,ZN15} (see \autoref{subsection:community_detection} for more details).
In consequence, the efficiency of the unlearning process will be affected.
Indeed, a revoked record would be more likely to belong to a large shard whose retraining time would be larger.
Therefore, in the context of GNNs, the unlearning algorithm should satisfy the following objectives:
\begin{itemize}[leftmargin=*]
    \item \mypara{G1: Balanced Shards}
    Different shards should share a similar size in terms of the number of nodes in each shard.
    In this way, each shard's retraining time is similar, which improves the efficiency of the whole graph unlearning process.
    Enforcing this objective can automatically satisfy the general unlearning pursuit of high unlearning efficiency.
    \item \mypara{G2: Comparable Model Utility}
    Graph structural information is the major factor that determines the performance of GNN~\cite{KW17, HYL17, WHX19}.
    To achieve comparable model utility, i.e., high prediction accuracy in node classification tasks, each shard should preserve the structural properties of the training graph.
\end{itemize}

\subsection{\sysname Framework}
\label{subsection:unlearning_pipeline}
To address the above mentioned challenges of unlearning in GNNs, we propose \sysname, which consists of the following three phases: Balanced graph partition, shard model training, and shard model aggregation.
The general framework of \sysname is illustrated in \autoref{figure:pipeline}.

\mypara{Balanced Graph Partition}
It is a crucial step of \sysname to fulfill the two requirements defined in \autoref{subsection:problem_definition}.
We propose to use balanced graph partition methods to partition the training graph into disjoint shards.
We discuss the existing balanced graph partition methods in \autoref{section:related}, and explain our proposed two balanced graph partition methods in \autoref{section:graph_partition}.

\mypara{Shard Model Training}
After the training graph is partitioned, the model owner can train one model for each of the shard graph, referred to as the \textit{shard model} ($\model_{i}$).
All shard models share the same model architecture.
To further speed up the training process, the model owner can train isolated shard models in parallel.

\mypara{Shard Model Aggregation}
At the inference phase, for predicting the label of node $w$, \sysname sends the corresponding data (the features of $w$, the features of its neighbors, and the graph structure among them) to all the shard models simultaneously, and the final prediction is obtained by aggregating the predictions from all the shard models.
We discuss the existing aggregation strategies and introduce our learning-based aggregation \optaggr in \autoref{section:optimal_aggregation}.

\begin{figure}[!t]
    \centering
    \begin{subfigure}{0.49\columnwidth}
    \includegraphics[width=\textwidth]{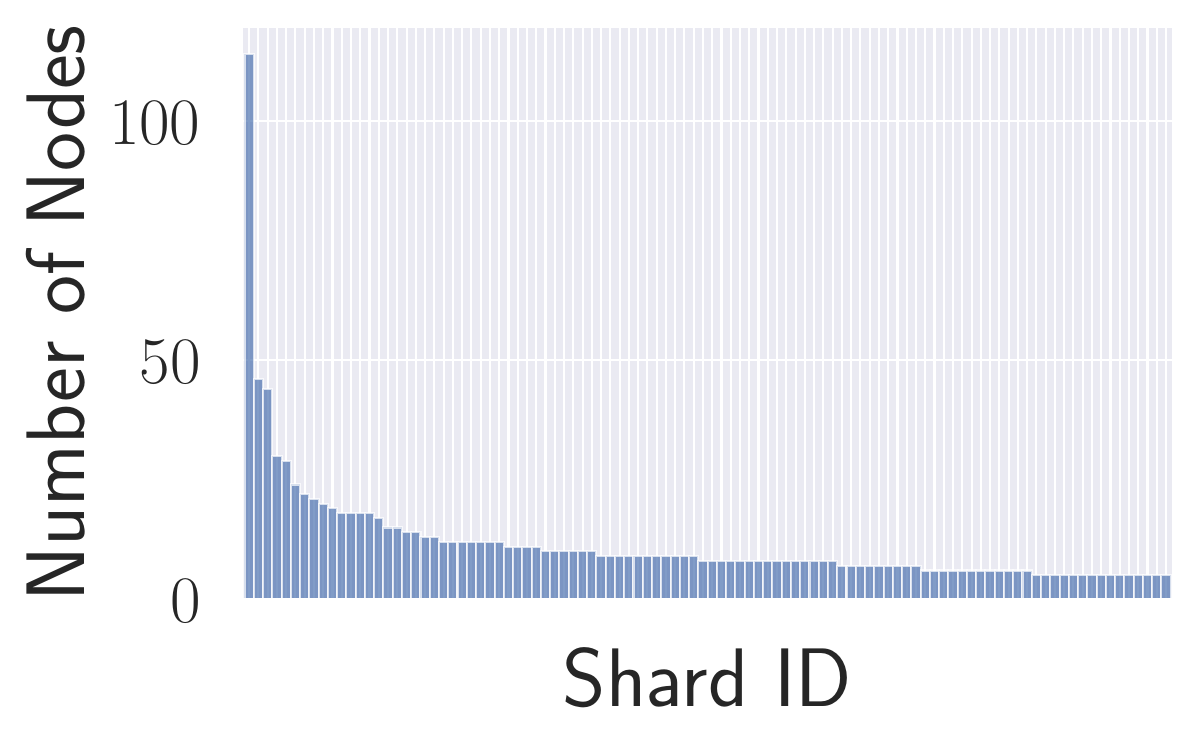}
    \caption{LPA (Top 100 shards)}
    \label{subfigure:lpa_cora_unbalance}
    \end{subfigure}
    \begin{subfigure}{0.49\columnwidth}
    \includegraphics[width=\textwidth]{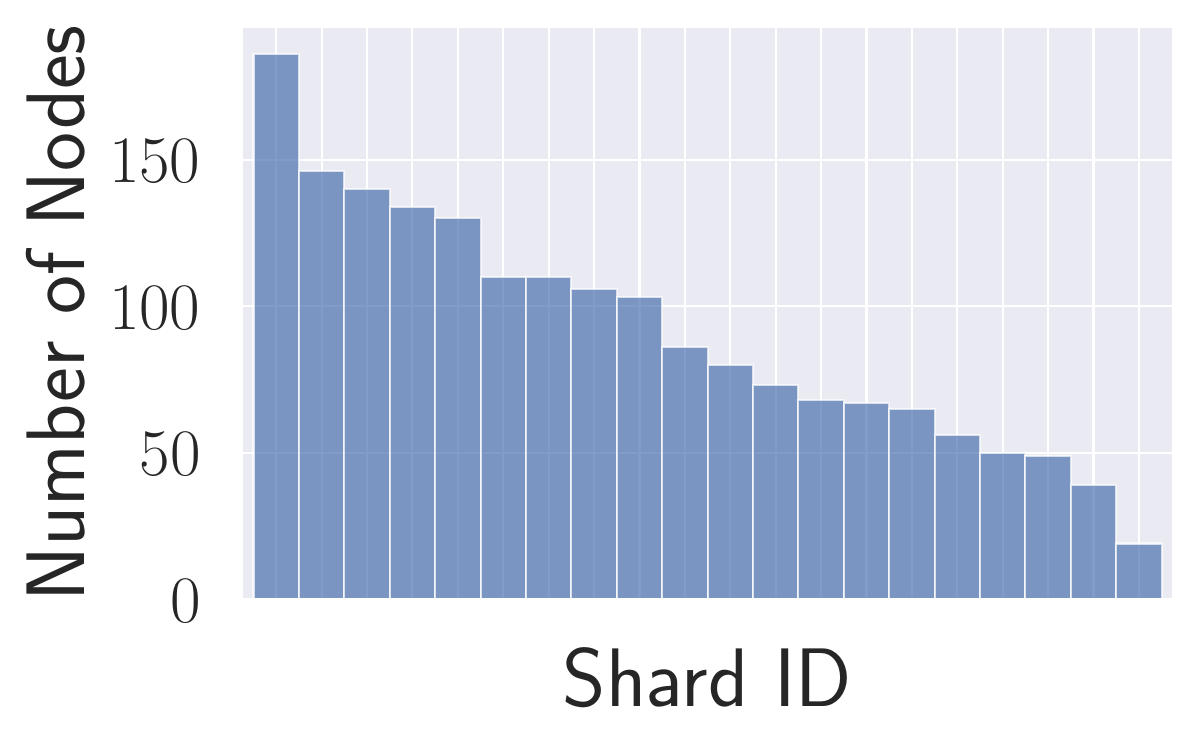}
    \caption{Embedding clustering}
    \label{subfigure:km_cora_unbalance}
    \end{subfigure} \\ %
    \vspace{-0.3cm}
    \caption{Distribution of shard sizes with classical graph partition methods on the Cora dataset.}
    \label{figure:shard_size_distribution}
\end{figure}

\section{Balanced Graph Partition}
\label{section:graph_partition}
In this section, we introduce the graph partition module.
Considering the node features and graph structural information in graph data, we identify three graph partition strategies.

\begin{itemize}[leftmargin=*]
    \item \mypara{Strategy 0}
    Consider the node feature information only and randomly partition the nodes.
    Concretely, we assume the node features are independently and identically distributed as in \sisa. In this sense, we can randomly partition the graph based on its node IDs.
\end{itemize}

\noindent This strategy can perfectly satisfy \textbf{G1} (Balanced Shards) in \autoref{subsection:problem_definition}, while it cannot satisfy \textbf{G2} (Comparable Model Utility) since it can destroy the structural information of the graph.
Thus, we treat this strategy as a baseline strategy.
To address \textbf{G2}, we also propose \textbf{Strategy 1} and \textbf{Strategy 2}.
\begin{itemize}[leftmargin=*]
    \item \mypara{Strategy 1}
    Consider the structural information only and try to preserve it as much as possible.
    One promising approach to do this is relying on community detection~\cite{WZ08, WL20a}.
    
    \item \mypara{Strategy 2}
    Consider both the structural information and the node features.
    To implement this, we can first represent the node features and graph structure into low-dimensional vectors, namely node embeddings, and then cluster the node embeddings into different shards.
\end{itemize}

\noindent However, directly applying them can result in a highly unbalanced graph partition due to the underlying structural properties of real-world graphs (see the distribution of shard sizes with classical partition methods in \autoref{figure:shard_size_distribution}).
To address this issue, we propose a general principle for achieving balanced graph partition (corresponding to \textbf{G1}), and apply this principle to design new approaches to achieve balanced graph partition for both \textbf{Strategy 1} and \textbf{Strategy 2}.
In the following, we elaborate on our balanced graph partition algorithms for \textbf{Strategy 1} and \textbf{Strategy 2}.

\subsection{Community Detection Method}
\label{subsection:community_detection}
For \textbf{Strategy 1}, we rely on community detection, which aims at dividing the graph into groups of nodes with dense connections internally and sparse connections between groups.
A spectrum of community detection methods have been proposed, such as Louvain~\cite{ZN15}, Infomap~\cite{RB08}, and Label Propagation Algorithm (LPA)~\cite{RAK07,WZ08}.
Among them, LPA has the advantage of low computational overhead and superior performance.
Thus, in this paper, we rely on LPA to design our graph partition algorithm.
For consistency purposes, we use \textit{shard} to represent the \textit{community}.

\begin{figure}[!t]
\centering
    \begin{subfigure}{0.3\columnwidth}
    \includegraphics[width=\columnwidth]{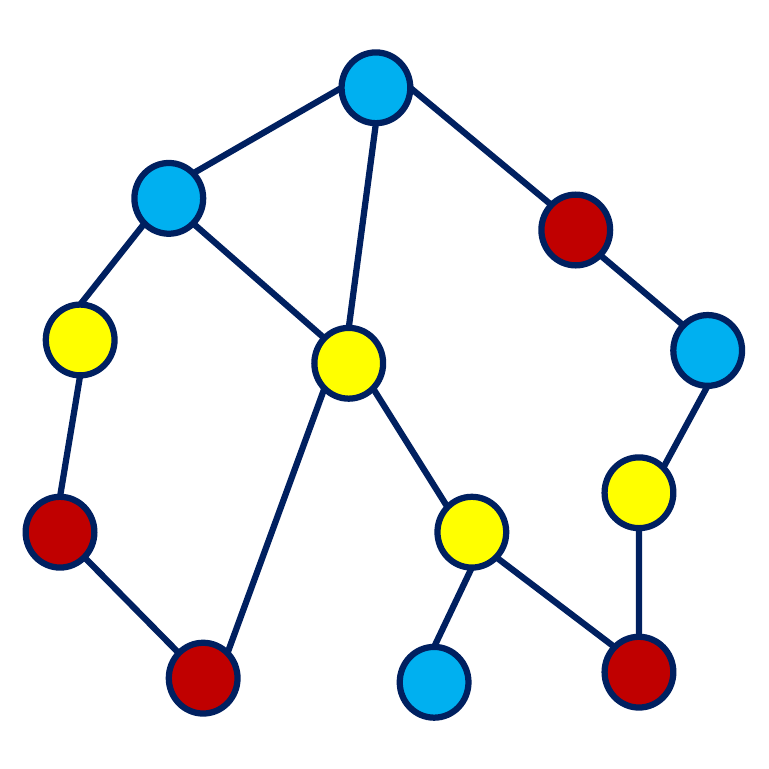}
    \caption{Initial state}
    \label{subfigure:lpa1}
    \end{subfigure}
    \begin{subfigure}{0.3\columnwidth}
    \includegraphics[width=\columnwidth]{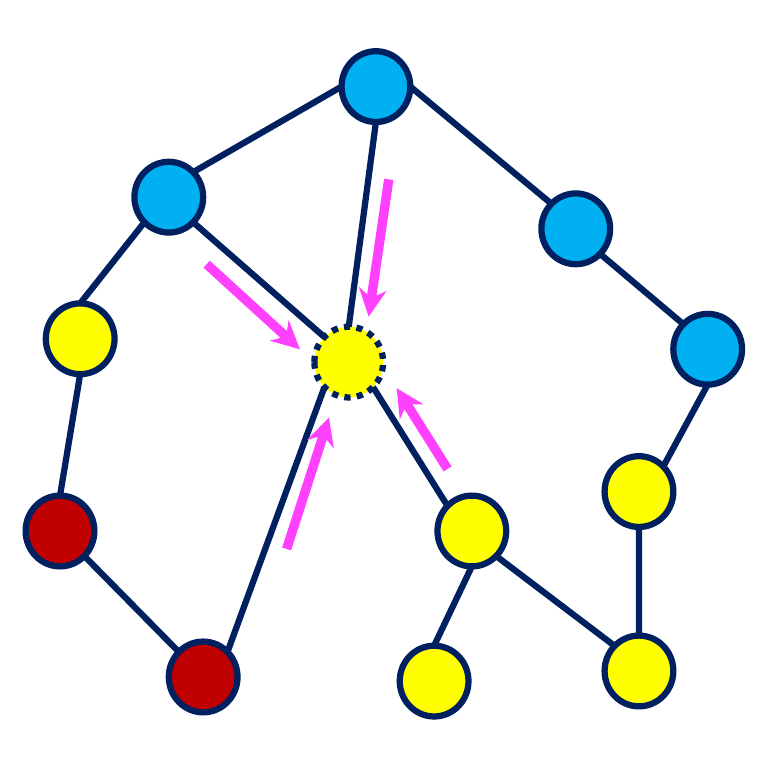}
    \caption{Propagation}
    \label{subfigure:lpa2}
    \end{subfigure}
    \begin{subfigure}{0.3\columnwidth}
    \includegraphics[width=\columnwidth]{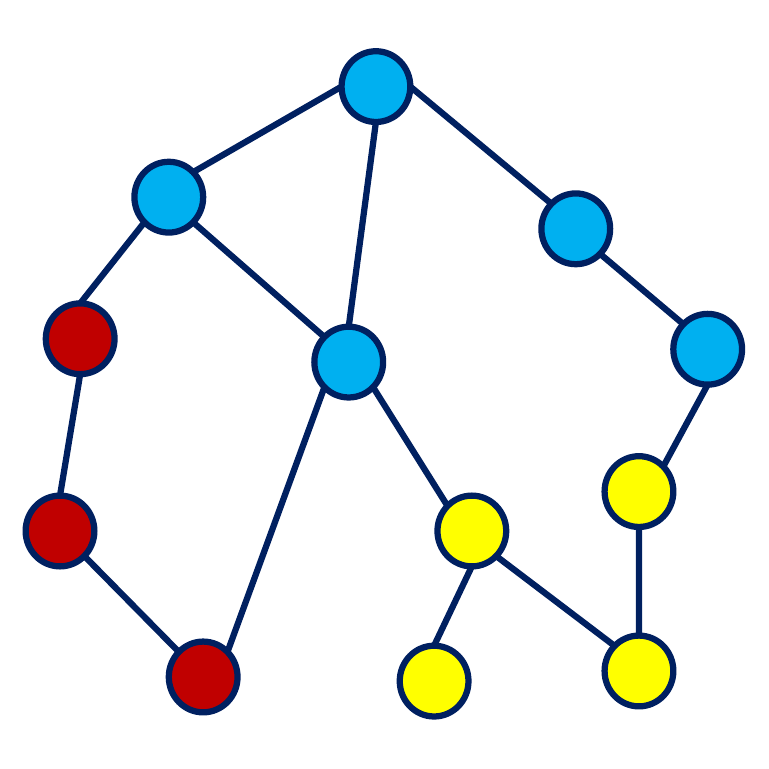}
    \caption{Stable state}
    \label{subfigure:lpa3}
    \end{subfigure}
    \vspace{-0.3cm}
    \caption{
    Illustration of LPA's workflow. 
    Different colors represent different shards.
    }
    \label{figure:lpa}
\end{figure} 

\mypara{Label Propagation Algorithm}
\autoref{figure:lpa} gives an illustration of the workflow of LPA.
At the initial state, each node is assigned a random shard label (\autoref{subfigure:lpa1}).
During the label propagation phase (\autoref{subfigure:lpa2} $\rightarrow$ \autoref{subfigure:lpa3}), each node sends out its own label, and updates its label to be the majority of the labels received from its neighbors.
For instance, the yellow node with a dashed outline in \autoref{subfigure:lpa2} will change its label to blue because the majority of its neighbors (two nodes above it) are labeled blue.
The label propagation process iterates through all nodes multiple times until convergence (there are no nodes changing their labels).

\mypara{Unbalanced Partition}
LPA is intriguing and powerful; however, directly applying the classical LPA results in a highly unbalanced graph partition.
For instance, \autoref{subfigure:lpa_cora_unbalance} shows the distribution of shard size on the Cora dataset~\cite{YCS16} ($2166$ nodes in the training graph).
We observe that the largest shard contains $113$ nodes, while the smallest one contains only $2$ nodes.
We provide a visualization of the shards detected by classical LPA in \autoref{app:visualization_lpa}.
Directly adopting the unbalanced shards detected by the classical LPA does not satisfy \textbf{G1}, which severely affects the unlearning efficiency.
For instance, if the revoked node is in the largest shard, there is not much benefit in terms of unlearning time.

\mypara{General Principle to Achieve Balanced Partition}
To address the above issue, we propose a general recipe to achieve balanced graph partition.
Given the desired shard number $k$ and maximal shard size $\delta$, we define a \textit{preference} for every \textit{node-shard pair} representing the node is assigned to the shard (which is referred to as \textit{destination shard}).
This results in $k \times n$ node-shard pairs with different preference values.
Then, we sort the node-shard pairs by preference values.
For each pair in descending preference order, we assign the node to the destination shard if the current number of nodes in that destination shard does not exceed $\delta$.

\mypara{Balanced LPA (\blpa)}
Following the general principle for achieving balanced partition, we define the preference as the \textit{neighbor counts} (the number of neighbors belonging to a destination shard) of the node-shard pairs, and the node-shard pairs with larger neighbor counts have higher priority to be assigned.

\autoref{alg:constrained_lpa} gives the workflow of \blpa.
The algorithm takes as input the set of nodes \nodeset, the adjacency matrix \adj, the number of desired shards $k$, the maximum number of nodes in each shard $\delta$, maximum iteration $T$, and works in four steps as follows:
\begin{itemize}[leftmargin=*]
    \item \mypara{Step 1: Initialization}
    We first randomly assign each node to one of the $k$ shards (\autoref{algline:blpa_init}).

    \item \mypara{Step 2: Reassignment Profiles Calculation}
    For each node $u$,
    we denote its \textit{reassignment profile} using a tuple $\tuple{u, \community_{src}, \community_{dst}, \neighcount}$, where $\community_{src}$ and $\community_{dst}$ are the current and destination shards of node $u$, $\neighcount$ is the neighbor counts of the destination shard $\community_{dst}$ (\autoref{algline:blpa_cal_profile_start} - \autoref{algline:blpa_cal_profile_end}).
    We store all the reassignment profiles in \comreassign.
    
    \item \mypara{Step 3: Reassignment Profiles Sorting}
    We rely on the intuition that the reassignment profile with larger neighbor counts should have a higher priority to be fulfilled; thus we sort \comreassign in descending order by \neighcount and obtain $\comreassign_s$ (\autoref{algline:blpa_sort_profile}).
    
    \item \mypara{Step 4: Label Propagation}
    Finally, we enumerate every instance of $\comreassign_s$.
    If the size of the destination shard $\community_{dst}$ does not exceed the given threshold $\delta$, we add the node $u$ to the destination shard and remove it from the current shard (\autoref{algline:blpa_propagation_start} - \autoref{algline:blpa_propagation_end}).
    After that, we remove all the remaining tuples containing node $u$ from $\comreassign_s$.
\end{itemize}

\noindent The \blpa algorithm repeats steps 2-4 until it reaches the maximum iteration $T$, or the shard does not change (\autoref{algline:blpa_terminate_start} - \autoref{algline:blpa_terminate_end}).

\mypara{Algorithm Analysis}
The computational complexity of \blpa depends on the size of the reassignment profile \comreassign.
Based on its definition, the number of tuples of each node $u$ in \comreassign equals to the number of neighbors of $u$.
Thus, the computational complexity of \blpa is $\mathcal{O}(n\cdot d_{ave})$, where $n$ is the number of nodes, and $d_{ave}$ is the average node degree of the training graph.

Regarding the convergence of \blpa, it is difficult to theoretically prove the convergence. 
Instead, we conduct empirical experiments to validate the convergence performance by checking the number of changed nodes in each iteration.
The experimental results show that the ratio of moved nodes gradually approximates to zero within 30 iterations on all five datasets.
Therefore, we set the number of iterations $T$ to 30 for all our experiments.
We refer the readers to \autoref{app:convergence} for the detailed experimental results.

\begin{algorithm}[!t]
\SetCommentSty{small}
\LinesNumbered
\caption{\blpa Algorithm}
\label{alg:constrained_lpa}

\KwIn{The set of all nodes \nodeset,
adjacency matrix \adj,
number of shards $k$,
maximum number of nodes in each shard $\delta$;
maximum iteration $T$;
}
\KwOut{Shards $\community = \{\community_1, \community_2, \cdots, \community_k\}$;}

\textbf{Initialization:}

Randomly allocate all nodes into $k$ shards and obtain $\community^0 = \{\community_1^0, \community_2^0, \cdots, \community_k^0\}$,
step $t=0$;
\label{algline:blpa_init}

\textbf{Label Propagation:}

\While{True}
{
    \ForEach{node $u$ in \nodeset}
    {
        \label{algline:blpa_cal_profile_start}
        \ForEach{shard $\community_{dst}$ in $\{ \community_i | v \in \neigh{u}, v \in \community_i \}$ }
        {
            Store tuple $\tuple{u, \community_{src}, \community_{dst}, \neighcount}$ in $\comreassign$;
            \label{algline:blpa_cal_profile_end}
        }
    }
        
    Sort \comreassign by \neighcount in descending order and obtain $\comreassign_s$;
    \label{algline:blpa_sort_profile}
    
    \ForEach{tuple in $\comreassign_s$}
    {
        \label{algline:blpa_propagation_start}
        \If{$|\community_{dst}^t| < \delta$}
        {
            $\community_{dst}^t \leftarrow \community_{dst}^{t-1} \cup u$; \\
            $\community_{src}^t \leftarrow \community_{src}^{t-1} \setminus u$;
            \label{algline:blpa_propagation_end} \\
            Remove all the remaining tuples containing node $u$ from $\comreassign_s$;
        }
    }
    \If{$t > T$ or the shard does not change}
    {
        \label{algline:blpa_terminate_start}
        break;
        \label{algline:blpa_terminate_end}
    }
    
    $t \leftarrow t + 1$;
}

\Return{$\community^t$}.
\end{algorithm}

\subsection{Embedding Clustering Method}
\label{subsection:embedding_clustering}

For \textbf{Strategy 2}, we rely on embedding clustering, which considers both the node feature and graph structural information for the partitioning.
Specifically, we first use a pretrained GNN model to obtain all the node embeddings, and then we perform clustering on the resulting node embeddings.

\mypara{Embedding Clustering}
We can adopt any state-of-the-art GNN models introduced in \autoref{subsection:gnn} to project each node into an embedding space.
With respect to clustering, we rely on the widely used $k$-means algorithm\cite{KMNPSW02}, which consists of three phases: Initialization, nodes reassignment, and centroids updating.
In the initialization phase, we randomly sample $k$ \textit{centroids} which represent the ``center'' of each shard.
In the node reassignment phase, each node is assigned to its ``nearest'' shard in terms of the Euclidean distance from the centroids.
In the centroids updating phase, the new centroids are recalculated as the average of all the nodes in their corresponding shard.

Similar to the case of the LPA method, directly using $k$-means can also produce highly unbalanced shards.
In \autoref{subfigure:km_cora_unbalance}, we observe that on the Cora dataset, the largest shard contains $10.24\%$ of the nodes, while the smallest one only contains $1.05\%$ of the nodes.

\mypara{Balanced Embedding $k$-means (\bkm)}
Following the same principle for achieving a balanced partition, we propose \bkm as shown in \autoref{alg:constrained_kmeans}.
We define the preference as the Euclidean distance between the node embedding and the centroid of the shard for all the node-shard pairs. 
A shorter distance implies a higher priority.
\bkm takes as input the set of all node embeddings $\embedset = \{\embed_1, \embed_2, \cdots, \embed_n\}$, the number of desired shards $k$, the maximum number of node embeddings in each shard $\delta$, the maximum number of iterations $T$, and works in four steps as follows:
\begin{itemize}[leftmargin=*]
    \item \mypara{Step 1: Initialization}
    We first randomly select $k$ centroids $\centroid^0 = \{\centroid_1^0, \centroid_2^0, \cdots, \centroid_k^0\}$ (\autoref{algline:bkm_init}).
    
    \item \mypara{Step 2: Embedding-Centroid Distance Calculation}
    Then, we calculate all the pairwise distance between the node embeddings and the centroids, which results in $n \times k$ embedding-centroid pairs.
    These pairs are stored in $\comreassign$ (\autoref{algline:bkm_cal_distance_start} - \autoref{algline:bkm_cal_distance_end}).
    
    \item \mypara{Step 3: Embedding-Centroid Distance Sorting}
    We rely on the intuition that the embedding-centroid pairs with closer distance have higher priorities; thus we sort $\comreassign$ in ascending order and obtain $\comreassign_s$ (\autoref{algline:bkm_sort_distance}).
    
    \item \mypara{Step 4: Node Reassignment and Centroid Updating}
    For each embedding-centroid pair in $\comreassign_s$, if the size of $\community_j$ is smaller than $\delta$, we assign node $u$ to shard $\community_j$ (\autoref{algline:bkm_reassign_start} - \autoref{algline:bkm_reassgin_end}).
    After that, we remove all the remaining tuples containing node $i$ from $\comreassign_s$.
    Finally, the new centroids are calculated as the average of all the nodes in their corresponding shards.
\end{itemize}

The \bkm algorithm repeats steps 2-4 until it reaches the maximum iteration $T$, or the centroid does not change (\autoref{algline:bkm_terminate_start} - \autoref{algline:bkm_terminate_end}).

\mypara{Algorithm Analysis}
Similar to \blpa, the computational complexity of \bkm depends on the size of \comreassign.
Since there are $n$ nodes and $k$ shards, the computational complexity of \bkm is $\mathcal{O}(k\cdot n)$.
We also empirically validate the convergence performance of \bkm. %

\subsection{Discussion}

\mypara{Choice of Graph Partition Algorithms}
The choice between \blpa and \bkm depends on the GNN structure.
In \autoref{subsection:model_utility_evaluation}, we provide a guideline on which one to choose.
In addition, we emphasize that \sysname is a general framework for graph unlearning, and any other balanced graph partition methods can be plugged into it.
In \autoref{subsection:comparison_balanced_partition}, we empirically compare our proposed \blpa and \bkm with several existing representative balanced graph partition methods, and show that our proposed methods are either more computational efficient or better performing.

\mypara{Guarantee of Unlearning}
Note that the shard models are deterministically unlearned but the clustering (graph partition) is not; thus, \sysname is doing approximate unlearning.
As such, we empirically quantify the possible information leakage using the state-of-the-art information leakage quantification method for machine unlearning system~\cite{CZWBHZ21} in \autoref{subsec:unlearning_power}, and show that \sysname does not leak much extra information.

Furthermore, from a legal-scholarship perspective, does not re-partition the graph also satisfies the right to be forgotten.
Note that legal-scholarship is vague and open to different explanations; below we just explain our understanding.
The third item of Art. 7 in the GDPR states that: ``\textit{The data subject shall have the right to withdraw his or her consent at any time. 
The withdrawal of consent \textbf{shall not affect the lawfulness of processing based on consent before its withdrawal}.}''
In our case, graph partition is a preprocessing step of the graph dataset, and the partitioned graph can be regarded as another form of the raw training graph.
It suffices to delete the data owners' revoked data from the processed data, e.g., remove the revoked node from the partitioned graph in our case, instead of removing the revoked data from the raw dataset and redo the preprocess operations.
This is supported by the application of the right to be forgotten in search engines~\cite{BBCCFFGHHIDKLNNOPSTV19}:
When the data owners ask to delete their data from the search results of a search engine, the service providers such as Google only need to directly delete the data from the current search results, instead of rerunning the ranking and recommendation algorithms on the raw data.

\begin{algorithm}[!t]
\SetCommentSty{small}
\LinesNumbered
\caption{\bkm Algorithm}
\label{alg:constrained_kmeans}

\KwIn{Node embeddings $\embedset = \{\embed_1, \embed_2, \cdots, \embed_n\}$, 
the number of clusters $k$,
maximum number of nodes embedding in each cluster $\delta$;
maximum number of iteration $T$;
}
\KwOut{Clusters $\community = \{\community_1, \community_2, \cdots, \community_k$\};}

\textbf{Initialization:}

Randomly select $k$ centroids $\centroid^0 = \{\centroid_1^0, \centroid_2^0, \cdots, \centroid_k^0\}$,
step $t = 0$;
\label{algline:bkm_init}

\While{True}
{
    \textbf{Nodes Reassignment:}
    
    \ForEach{node embedding $i \in \embedset$}
    {
        \label{algline:bkm_cal_distance_start}
        \ForEach{centroid $j \in \community$}
        {
            Store $||\embed_i - \centroid_j||_2$ in \comreassign;
            \label{algline:bkm_cal_distance_end}
        }
    }
    
    Sort \comreassign in ascending order and obtain $\comreassign_s$.
    \label{algline:bkm_sort_distance}
    
    \ForEach{node $i$ and centroid $j$ in $\comreassign_s$}
    {
        \label{algline:bkm_reassign_start}
        \If{$|\community_j^t| < \delta$}
        {
            $\community_j^t \leftarrow \community_j^t \cup i$; \\
            Remove all the remaining tuples containing node $i$ from $\comreassign_s$;
        }
    }
    
    \textbf{Centroids Updating:}
    
    \ForEach{cluster $j \in \community^t$}
    {
        $\centroid_j^t = \frac{\sum_{i \in \community_j^t} \embed_i}{|\community_j^t|}$;
        \label{algline:bkm_reassgin_end}
    }
    
    \If{$t > T$ or the centroid do not change}
    {
        \label{algline:bkm_terminate_start}
        break;
        \label{algline:bkm_terminate_end}
    }
    
    $t \leftarrow t + 1$;
}

\Return{$\community^t$}.
\end{algorithm}

\section{Learning-based Aggregation \\(\optaggr)}
\label{section:optimal_aggregation}
\mypara{Existing Aggregation Strategies}
The most straightforward aggregation strategy, also mainly used in~\cite{BCCJTZLP21}, is majority voting, where each shard model predicts a label and $w$ takes the label predicted most often. 
We refer to this aggregation strategy as \majaggr.
An alternative solution is to gather the posterior vectors of all shard models and average them to obtain aggregated posteriors.
The target nodes are predicted as the highest posterior in this aggregation.
We refer to this aggregation strategy as \meanaggr.

Note that different shard models can have different contributions to the final prediction; thus allocating the same \textit{importance score} for all shard models during the aggregation phase might not achieve the best prediction accuracy. 

\mypara{Our Proposal}
In this section, we propose a learning-based aggregation method \optaggr.
We assign an importance score to each shard model, which can be learned based on the following loss function.
\begin{align}
    \underset{\alpha}{\min} \; \underset{w \in \graph_o}{\mathbb{E}}\left[ \mathcal{L} \left( \sum_{i=0}^{m} \alpha_{i} \cdot \model_{i}(\feat_w, \neigh{w}), y \right) \right]
    + \lambda\sum_{i=0}^{m} \vert\vert \alpha_{i} \vert\vert \label{equation:optaggr}
\end{align}
where $\feat_w$ and $\neigh{w}$ are the feature vector and neighborhood of a node $w$ from the training graph, $y$ is the true label of $w$, $\model_i(\cdot)$ represents shard model $i$, $\alpha_i$ is the importance score for $\model_i(\cdot)$, and $m$ is the total number of shards.
We regulate the summation of all importance scores to $1$.
Further, \loss represents the loss function and we adopt the standard cross-entropy loss in this paper.
The regularization term $\vert\vert \cdot \vert\vert$ is used to reduce overfitting.

\mypara{Solving the Optimization Problem}
We can run gradient descent to find the optimal $\alpha$ to solve the optimization problem.
However, directly running gradient descent can result in negative values in $\alpha$.
To address this issue, after each gradient descent iteration, we map the negative importance score back to $0$.
The mapping the negative importance scores to 0 follows the general idea of projected gradient descent (PGD)~\cite{APY19}.
In addition, the summation of the importance scores could deviate from $1$.
We first tried to normalize the importance score using the summation of current scores in each iteration; however, we empirically found that the loss could be extremely unstable across different epochs.
Thus, we instead use the softmax function for normalization in each iteration.

\begin{algorithm}[!t]
\SetCommentSty{small}
\LinesNumbered
\caption{\sysname}
\label{alg:overall_workflow}

    \KwIn{Training graph $\graph_0$,
    GNN model type $f$,
    number of shards $k$,
    maximum number of nodes in each shard $\delta$,
    maximum iteration $T$;
    }
    
    \KwOut{
    Shard models $\model = \{\model_1, \model_2, \cdots, \model_k\}$,
    importance scores $\alpha = \{\alpha_1, \alpha_2, \cdots \alpha_k\}$;
    }
    
    \textbf{Graph Partition:}
    \label{algline:graph_partition_start}
    
    \If{the GNN model type $f$ is \gcn:}
    {
        Partitioning $\graph_0$ into $k$ shards with \autoref{alg:constrained_lpa} and obtain $\graph_s = \{\graph_s^1, \graph_s^2, \cdots, \graph_s^k$\};
    }
    
    \Else
    {
        Partitioning $\graph_0$ into $k$ shards with \autoref{alg:constrained_kmeans} and obtain $\graph_s = \{\graph_s^1, \graph_s^2, \cdots, \graph_s^k$\};
        \label{algline:graph_partition_end}
    }
    
    \textbf{Shard Model Training:}
    \label{algline:shard_model_train_start}
    
    Using $\graph_s$ to train shard models $\model = \{\model_1, \model_2, \cdots, \model_k\}$;
    \label{algline:shard_model_train_end}
    
    \textbf{Importance Scores Learning:}
    \label{algline:importance_score_learn_start}
    
    Randomly sampling a set of nodes $\nodeset_0$ from $\graph_0$;
    
    Replacing $\graph_0$ in \autoref{equation:optaggr} with $\nodeset_0$ and train $\alpha$;
    \label{algline:importance_score_learn_end}
    
    \Return{$\model$, $\alpha$}.
\end{algorithm}

\mypara{Importance Scores {Unlearning}}
{
Note that the nodes that learn the optimal importance scores can also be revoked by their data subjects.
Therefore, we need to relearn the shard importance scores if a request-to-unlearn node is used to train the \optaggr, and this learning time is counted as part of the unlearning time.
To reduce this relearning time, we propose to use only a small random subset of nodes from the training graph to relearn.
We empirically show in \autoref{subsection:optaggr_effectiveness} that using only 10\% of the nodes in the training graph can achieve comparable utility as using all nodes. 
In this sense, relearning the optimal shard importance scores is unnecessary when the unlearned nodes are not used to train the \optaggr.
}

\subsection{Putting Things Together: \sysname}
\autoref{alg:overall_workflow} illustrates the overall workflow of \sysname.
It takes as input the training graph $\graph_0$, the GNN model type $f$, and all necessary parameters for \autoref{alg:constrained_lpa} and \autoref{alg:constrained_kmeans} ($k$, $\delta$, and $T$).
If $f$ is a \gcn, we invoke \autoref{alg:constrained_lpa} to partition $\graph_0$; otherwise, we use \autoref{alg:constrained_kmeans} (\autoref{algline:graph_partition_start} - \autoref{algline:graph_partition_end}).
We then use the partitioned graph $\graph_s$ to train a set of shard models $\model$ (\autoref{algline:shard_model_train_start}).
Finally, we randomly sample a set of nodes $\nodeset_0$ from $\graph_0$ to train the importance scores $\alpha$ for each shard models.
The shard models and importance scores produced by \sysname can be used to predict the label of new samples.
When some nodes or edges are revoked by the data owner, we only need to retrain the corresponding shard model.

\section{Evaluation}
\label{section:experiment}
In this section, we first evaluate the unlearning efficiency and model utility of \sysname, respectively.
Second, we conduct experiments to show the superiority of our proposed learning-based aggregation method \optaggr.
Third, we compare our proposed balanced graph partition methods with existing methods.
Fourth, we illustrate the unlearning power of \sysname.

In addition, we investigate the following issues, and due to space limitation, the corresponding results are deferred to the appendix:
(1) We investigate the correlation between the properties of the shard models and the importance scores resulting from \optaggr (\autoref{app:shard_importance}).
(2) We investigate the impact of graph structure in a more controllable manner (\autoref{app:role_graph_structure}).
(3) We show the robustness of \sysname to the number of unlearned nodes/edges (\autoref{app:unlearning_robustness}).
(4) We conduct ablation studies to show the impact of $k$ and $\delta$ on the unlearning efficiency and model utility (\autoref{app:ablation}).

\subsection{Experimental Setup}

\mypara{Datasets} 
We conduct our experiments on five widely used public graph datasets~\cite{KW17, SHFZWS20a}, including Cora, Citeseer, Pubmed~\cite{YCS16}, CS~\cite{SMBG18a}, and Physics~\cite{SMBG18a}. 
\autoref{table:datasets} summarizes the statistics of all the datasets.
Cora, Citeseer, and Pubmed are citation datasets, where the nodes represent the publications, and there is an edge between two publications if one cites the other.
The node features are binary vectors indicating the presence of the keywords from a dictionary, and the class labels represent the publications' research field. 
CS and Physics are coauthor datasets, where two authors are connected if they collaborate on at least one paper.
The node features represent paper keywords for each author's papers, and the class labels indicate each author's most active fields of study.

\mypara{GNN Models}
We evaluate the efficiency and utility of \sysname on four state-of-the-art GNN models, including \sage, \gcn, \gat, and \gin, their details are discussed in \autoref{app:gnn_details}.
For each GNN model, we stack two layers of GNN modules.
All the models are implemented with the PyTorch Geometric\footnote{\url{https://github.com/rusty1s/pytorch_geometric}} library.
All the GNN models (including the shard models) considered in this paper are trained for $100$ epochs. We use Adam optimizer and set default learning rate to 0.01 with 0.001 weight decay.

\mypara{Metrics}
In the design of \sysname, we mainly consider two aspects of performance, unlearning efficiency and model utility.
\begin{itemize}[leftmargin=*]
    \item \mypara{Unlearning Efficiency}
    Directly measuring the unlearning time for one unlearning request is inaccurate due to the diversity of shards.
    Thus, we calculate the \textit{average unlearning time} for $100$ independent unlearning requests.
    Concretely, we randomly sample $100$ nodes/edges from the training graph, record the retraining time of their corresponding shard models, and calculate the average retraining time.
    
    \item \mypara{Model Utility}
    We use the \textit{Micro F1 score} to measure the model utility, which is widely used to evaluate the prediction ability of GNN models on multi-class classification~\cite{GL16}.
    The F1 score is a harmonic mean of \textit{precision} and \textit{recall}, and can provide a good measure of the incorrectly classified cases.
\end{itemize}

\begin{table}[t]
    \centering
    \caption{Dataset statistics.}
    \label{table:datasets}
    \footnotesize
    \resizebox{0.95\linewidth}{!}{
    \setlength{\tabcolsep}{0.5em}
    \begin{tabular}{c|c c c c c}
    \toprule
    \textbf{Dataset} & \textbf{Category} & \#.\ \textbf{Nodes} & \#.\ \textbf{Edges} & \#.\ \textbf{Classes} & \#.\ \textbf{Features} \\
    \toprule
    \textbf{Cora} & Citation  & 2,708 & 5,429 & 7 & 1,433 \\
    \rowcolor{mygray}
    \textbf{Citeseer} & Citation  & 3,327 & 4,732 & 6 & 3,703 \\
    \textbf{Pubmed} & Citation  & 19,717 & 44,338 & 3 & 500 \\
    \rowcolor{mygray}
    \textbf{CS} & Coauthor & 18,333 & 163,788 & 15 & 6805 \\
    \textbf{Physics} & Coauthor & 34,493 & 495,924 & 5 & 8415 \\
    \bottomrule
    \end{tabular}
    }
\end{table}

\mypara{Competitors}
We have two natural baselines: The training from scratch method (which is referred to as \scratch) and the random method (which is based on partitioning the training graph randomly rely on Strategy 0, and we refer it to as \random).
The \scratch method can achieve good model utility but its unlearning efficiency is low. 
On the other hand, the \random method can achieve high unlearning efficiency but suffers from poor model utility.

We implement both community detection and embedding clustering based graph partition methods in \autoref{section:graph_partition} for \sysname.
For presentation purpose, we refer them to as \sysname-\blpa and \sysname-\bkm, respectively.

\mypara{Experimental Settings}
For each dataset, we randomly split the whole graph into two disjoint parts, where $80\%$ of nodes are used in training GNN models, and $20\%$ of nodes are used to evaluate the model utility.
Note that the graph partition algorithms are only applied to the training graph.
By default, we set the number of shards $k$ for Cora, Citeseer, Pubmed, CS, and Physics to 20, 20, 50, 30, and 100, respectively, which ensures each shard is trained on a reasonable number of nodes and edges.
The maximum number of nodes in each shard $\delta$ is set as $\ceil{\frac{n}{k}}$.
We validate the effectiveness of this setting in \autoref{app:ablation}. 
The maximum number of iterations $T$ of both \blpa and \bkm are set to $30$ as we empirically showed that $T=30$ can guarantee the convergence of both algorithms.
Besides, we set the embedding dimension of \bkm as 32. 
Note that if the embedding dimension is large, the dimension reduction method might help.

\mypara{Implementation}
We implement \sysname with Python 3.7 and PyTorch 1.7.
All experiments are run on an NVIDIA DGX-A100 server with 2 TB memory and Ubuntu 18.04 LTS system.
All the experiments regarding to model utility are run 10 times and we report the mean and standard deviation.

\begin{figure*}[!t]
\begin{center}
\includegraphics[width=0.8\textwidth]{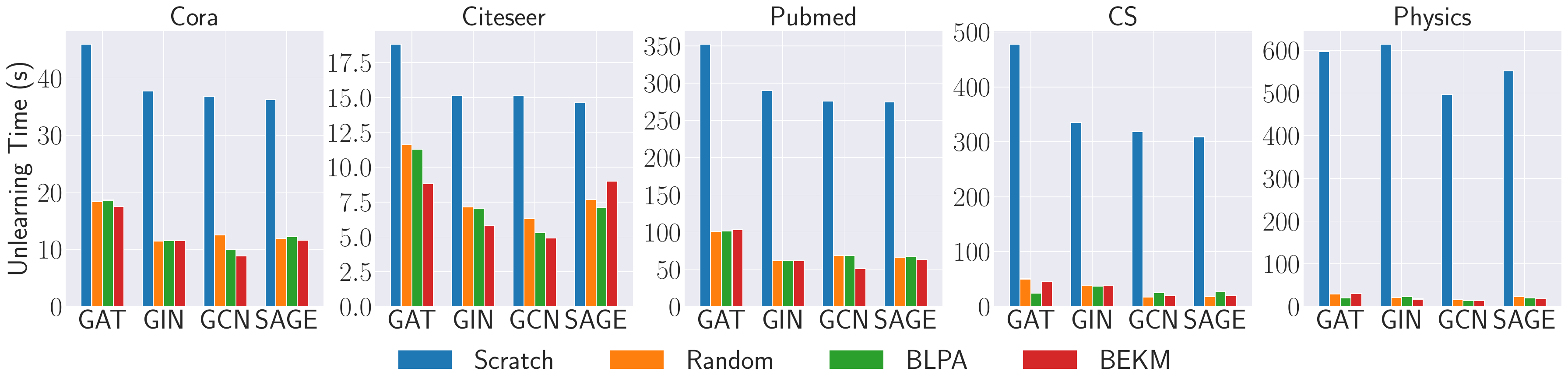}
\end{center}
\vspace{-0.3cm}
\caption{Comparison of node unlearning efficiency for all graph unlearning methods.
\blpa and \bkm stand for \sysname-\blpa and \sysname-\bkm unlearning methods, respectively. We omit the results of edge unlearning to avoid repetitive conclusions.
The results of edge unlearning are in \autoref{figure:edge_unlearning_efficiency}.
}
\label{figure:node_unlearning_efficiency}
\end{figure*}

\begin{table}
    \centering
    \caption{Computational costs of the \sysname pipeline on five datasets.
    We report the prediction cost and the relearning cost of \optaggr for \bkm.}
    \label{table:time_cost}
    \resizebox{1\linewidth}{!}{
    \footnotesize
    \setlength{\tabcolsep}{0.3em}
    \begin{tabular}{c | c c c | c c | c}
    \toprule
    \multicolumn{1}{c|}{} 
    & \multicolumn{3}{c|}{\textbf{Graph Partition Cost}}
    & \multicolumn{2}{c|}{\textbf{Prediction Cost}} 
    & \multicolumn{1}{c}{\textbf{\textbf{Learn Cost of}}} \\
    \textbf{Dataset} & \random & \blpa & \bkm  & \scratch & Shard & \optaggr \\
    \toprule
    \multirow{1}{*}{\rotatebox[origin=c]{0}{\textbf{Cora}}}
    & 0.8s & 3s & 26s & 0.002s & 0.003s & 1.3s \\
    \rowcolor{mygray}
    \multirow{1}{*}{\rotatebox[origin=c]{0}{\textbf{Citeseer}}}
    & 0.5s & 2s & 20s & 0.003s & 0.004s & 1.5s \\
    \multirow{1}{*}{\rotatebox[origin=c]{0}{\textbf{Pubmed}}}
    & 1s & 20s & 240s & 0.004s & 0.008s & 19s \\
    \rowcolor{mygray}
    \multirow{1}{*}{\rotatebox[origin=c]{0}{\textbf{CS}}}
    & 1s & 13s & 220s & 0.004s & 0.009s & 25s \\
    \multirow{1}{*}{\rotatebox[origin=c]{0}{\textbf{Physics}}}
    & 1s & 40s & 480s & 0.005s & 0.021s & 33s  \\
    \bottomrule
    \end{tabular}
    }
\end{table}

\subsection{Evaluation of Unlearning Efficiency}
\label{subsection:unlearning_efficiency_evaluation}
In this section, we evaluate the unlearning efficiency of different graph unlearning methods on five datasets and four GNN models. 

\mypara{Setup}
\autoref{figure:node_unlearning_efficiency} illustrates the node unlearning efficiency for different graph unlearning methods.
For the shard-based unlearning methods, i.e., \random, \sysname-\blpa, and \sysname-\bkm, each unlearning request time cost consists of two parts: Retraining the shard models and relearning the importance scores of \optaggr.
As discussed in \autoref{section:optimal_aggregation}, we only use a small portion of nodes in the training graph to learn the importance scores.
The \textit{average relearning time} of \optaggr on all datasets is shown in the last column of \autoref{table:time_cost}.
The results show that the relearning time is less than 30s for most of the datasets, which is negligible compared to retraining the shard models.

\mypara{Results}
We observe that the shard-based unlearning methods can significantly improve the unlearning efficiency compared to the \scratch method.
For all the four GNN models, we observe a similar time efficiency improvement level.
In addition, the relative efficiency improvement of larger datasets (Pubmed, CS, and Physics) is more significant than that of smaller datasets (Cora and Citeseer). 
For instance, the unlearning time improvement is of $4.16\times$ for the Cora dataset, $3.08\times$ for the Citeseer dataset, $5.40\times$ for the Pubmed dataset, $19.25\times$ for the CS dataset, and $35.9\times$ for the Physics datasets.
This is expected.
From the \scratch method perspective, training a large graph can cost a large amount of time.
From the shard-based methods perspective, we can tolerate more shards for larger graphs while preserving the model utility.
Comparing different shard-based methods, we observe that \sysname-\blpa and \sysname-\bkm have similar unlearning time as \random.
This is made possible by our approach for achieving balanced partition with \blpa and \bkm (see \autoref{section:graph_partition}).

\mypara{Additional Time Cost Analysis}
Besides the unlearning cost, there are two additional costs in the \sysname pipeline: Graph partition cost and prediction cost.
\autoref{table:time_cost} illustrates these two costs on five datasets.
We observe that the graph partition costs of \blpa and \bkm are higher than \random.
This is expected since both \blpa and \bkm need to iterate multiple times to preserve the structural information.
Once the graph partition is done, we keep it fixed without unlearning it.
In this sense, we can tolerate this cost since it is only executed once.
We empirically validate that using a fixed partition does not result in noticeable model utility degradation for \sysname in \autoref{app:unlearning_robustness}.

For the prediction cost, the shard-based methods are slightly more time-consuming compared to the \scratch method, since we need to obtain the prediction from all shard models and aggregate them.
Fortunately, the prediction cost is negligible since most of their values are smaller than 0.01 second.

\begin{table*}[!t]
    \centering
    \caption{Comparison of F1 scores for unlearning methods and different aggregation methods.
    Note that the \scratch method does not need aggregation.
    We highlight the \scratch method in the \colorbox{green!15}{green} ground and our proposed methods in the \colorbox{red!15}{red}.
    For each graph partition strategy, we highlight the best value in \textbf{\textcolor{black}{bold}}.
    and for each GNN model, we highlight the best value in \textbf{\textcolor{blue}{blue bold}}.
    We omit the results of edge unlearning due to similar conclusions.
    }
    \label{table:node_unlearning_utility}
    \resizebox{0.95\linewidth}{!}{
    \setlength{\tabcolsep}{0.5em}
    \begin{tabular}{c c | c | c c c | c c c | c c c }
    \toprule
    & \multicolumn{1}{c|}{\textbf{Dataset}/} & \multicolumn{1}{c|}{\scratch} & \multicolumn{3}{c|}{\random} & \multicolumn{3}{c|}{\sysname-\blpa} & \multicolumn{3}{c}{\sysname-\bkm}\\
    & \textbf{Model} &  & \meanaggr & \majaggr & \optaggr & \meanaggr & \majaggr & \optaggr & \meanaggr & \majaggr & \optaggr \\
    \toprule
    &GAT & \cellcolor{green!15}0.823 $\pm$ 0.006 & 0.649 $\pm$ 0.006 & 0.638 $\pm$ 0.010 & \cellcolor{red!15}\textbf{0.706 $\pm$ 0.004} & 0.356 $\pm$ 0.005 & 0.492 $\pm$ 0.009 & \cellcolor{red!15}\textbf{0.727 $\pm$ 0.009} & 0.672 $\pm$ 0.004 & 0.669 $\pm$ 0.012 & \cellcolor{red!15}\textbf{\textcolor{blue}{0.754 $\pm$ 0.009}} \\
    \rowcolor{mygray}
    \cellcolor{white}
    &GCN & \cellcolor{green!15}0.739 $\pm$ 0.003 & 0.337 $\pm$ 0.006 & 0.188 $\pm$ 0.004 & \cellcolor{red!15}\textbf{0.509 $\pm$ 0.009} & 0.590 $\pm$ 0.008 & 0.319 $\pm$ 0.007 & \cellcolor{red!15}\textbf{\textcolor{blue}{0.676 $\pm$ 0.004}} & 0.390 $\pm$ 0.011 & 0.247 $\pm$ 0.012 & \cellcolor{red!15}\textbf{0.493 $\pm$ 0.006} \\
    &GIN & \cellcolor{green!15}0.787 $\pm$ 0.013 & 0.760 $\pm$ 0.030 & 0.702 $\pm$ 0.033 & \cellcolor{red!15}\textbf{0.736 $\pm$ 0.021} & 0.681 $\pm$ 0.039 & 0.594 $\pm$ 0.028 & \cellcolor{red!15}\textbf{0.753 $\pm$ 0.015} & 0.758 $\pm$ 0.016 & 0.742 $\pm$ 0.031 & \cellcolor{red!15}\textbf{\textcolor{blue}{0.801 $\pm$ 0.018}} \\
    \rowcolor{mygray}
    \cellcolor{white}
    \multirow{-4}{*}{\rotatebox[origin=c]{90}{\textbf{Cora}}}
    &SAGE & \cellcolor{green!15}0.824 $\pm$ 0.004 & 0.583 $\pm$ 0.009 & 0.572 $\pm$ 0.012 & \cellcolor{red!15}\textbf{0.682 $\pm$ 0.013} & 0.354 $\pm$ 0.008 & 0.486 $\pm$ 0.012 & \cellcolor{red!15}\textbf{0.684 $\pm$ 0.014} & 0.673 $\pm$ 0.008 & 0.646 $\pm$ 0.010 & \cellcolor{red!15}\textbf{\textcolor{blue}{0.740 $\pm$ 0.013}} \\
    \midrule
    &GAT & \cellcolor{green!15}0.691 $\pm$ 0.015 & 0.502 $\pm$ 0.012 & 0.507 $\pm$ 0.016 & \cellcolor{red!15}\textbf{0.631 $\pm$ 0.015} & 0.504 $\pm$ 0.010 & 0.486 $\pm$ 0.009 & \cellcolor{red!15}\textbf{0.676 $\pm$ 0.004} & 0.744 $\pm$ 0.007 & 0.712 $\pm$ 0.010 & \cellcolor{red!15}\textbf{\textcolor{blue}{0.746 $\pm$ 0.006}} \\
    \rowcolor{mygray}
    \cellcolor{white}
    &GCN & \cellcolor{green!15}0.493 $\pm$ 0.006 & 0.263 $\pm$ 0.014 & 0.157 $\pm$ 0.011 & \cellcolor{red!15}\textbf{0.277 $\pm$ 0.009} & 0.372 $\pm$ 0.006 & 0.192 $\pm$ 0.006 & \cellcolor{red!15}\textbf{\textcolor{blue}{0.450 $\pm$ 0.006}} & 0.298 $\pm$ 0.005 & 0.129 $\pm$ 0.007 & \cellcolor{red!15}\textbf{0.332 $\pm$ 0.006} \\
    &GIN & \cellcolor{green!15}0.587 $\pm$ 0.031 & 0.611 $\pm$ 0.028 & 0.540 $\pm$ 0.056 & \cellcolor{red!15}\textbf{0.626 $\pm$ 0.022} & 0.451 $\pm$ 0.062 & 0.447 $\pm$ 0.032 & \cellcolor{red!15}\textbf{0.612 $\pm$ 0.026} & 0.725 $\pm$ 0.016 & 0.696 $\pm$ 0.014 & \cellcolor{red!15}\textbf{\textcolor{blue}{0.739 $\pm$ 0.020}} \\
    \rowcolor{mygray}
    \cellcolor{white}
    \multirow{-4}{*}{\rotatebox[origin=c]{90}{\textbf{Citeseer}}}
    &SAGE & \cellcolor{green!15}0.668 $\pm$ 0.013 & 0.519 $\pm$ 0.024 & 0.536 $\pm$ 0.026 & \cellcolor{red!15}\textbf{0.623 $\pm$ 0.014} & 0.447 $\pm$ 0.007 & 0.472 $\pm$ 0.024 & \cellcolor{red!15}\textbf{0.657 $\pm$ 0.012} & 0.708 $\pm$ 0.003 & 0.710 $\pm$ 0.007 & \cellcolor{red!15}\textbf{\textcolor{blue}{0.716 $\pm$ 0.007}} \\
    \midrule
    &GAT & \cellcolor{green!15}0.851 $\pm$ 0.004 & 0.852 $\pm$ 0.001 & 0.851 $\pm$ 0.002 & \cellcolor{red!15}\textbf{0.857 $\pm$ 0.002} & 0.843 $\pm$ 0.002 & 0.840 $\pm$ 0.002 & \cellcolor{red!15}\textbf{0.858 $\pm$ 0.003} & 0.853 $\pm$ 0.001 & 0.852 $\pm$ 0.001 & \cellcolor{red!15}\textbf{\textcolor{blue}{0.860 $\pm$ 0.003}} \\
    \rowcolor{mygray}
    \cellcolor{white}
    &GCN & \cellcolor{green!15}0.748 $\pm$ 0.017 & 0.484 $\pm$ 0.004 & 0.207 $\pm$ 0.000 & \cellcolor{red!15}\textbf{0.551 $\pm$ 0.005} & 0.644 $\pm$ 0.004 & 0.423 $\pm$ 0.011 & \cellcolor{red!15}\textbf{\textcolor{blue}{0.718 $\pm$ 0.010}} & 0.353 $\pm$ 0.003 & 0.207 $\pm$ 0.000 & \cellcolor{red!15}\textbf{0.482 $\pm$ 0.003} \\
    &GIN & \cellcolor{green!15}0.837 $\pm$ 0.015 & 0.854 $\pm$ 0.003 & 0.852 $\pm$ 0.003 & \cellcolor{red!15}\textbf{0.856 $\pm$ 0.003} & 0.849 $\pm$ 0.002 & 0.843 $\pm$ 0.002 & \cellcolor{red!15}\textbf{0.855 $\pm$ 0.004} & 0.859 $\pm$ 0.002 & 0.851 $\pm$ 0.010 & \cellcolor{red!15}\textbf{\textcolor{blue}{0.859 $\pm$ 0.003}} \\
    \rowcolor{mygray}
    \cellcolor{white}
    \multirow{-4}{*}{\rotatebox[origin=c]{90}{\textbf{Pubmed}}}
    &SAGE & \cellcolor{green!15}0.874 $\pm$ 0.003 & 0.854 $\pm$ 0.002 & 0.852 $\pm$ 0.003 & \cellcolor{red!15}\textbf{0.857 $\pm$ 0.002} & 0.841 $\pm$ 0.003 & 0.836 $\pm$ 0.003 & \cellcolor{red!15}\textbf{\textcolor{blue}{0.863 $\pm$ 0.002}} & 0.854 $\pm$ 0.002 & 0.852 $\pm$ 0.002 & \cellcolor{red!15}\textbf{0.862 $\pm$ 0.002} \\
    \midrule
    &GAT & \cellcolor{green!15}0.919 $\pm$ 0.004 & 0.880 $\pm$ 0.001 & 0.877 $\pm$ 0.001 & \cellcolor{red!15}\textbf{0.882 $\pm$ 0.000} & 0.664 $\pm$ 0.015 & 0.662 $\pm$ 0.009 & \cellcolor{red!15}\textbf{0.858 $\pm$ 0.004} & 0.885 $\pm$ 0.001 & 0.882 $\pm$ 0.003 & \cellcolor{red!15}\textbf{\textcolor{blue}{0.906 $\pm$ 0.002}} \\
    \rowcolor{mygray}
    \cellcolor{white}
    &GCN & \cellcolor{green!15}0.903 $\pm$ 0.006 & 0.644 $\pm$ 0.002 & 0.528 $\pm$ 0.001 & \cellcolor{red!15}\textbf{0.706 $\pm$ 0.008} & 0.658 $\pm$ 0.004 & 0.440 $\pm$ 0.003 & \cellcolor{red!15}\textbf{0.750 $\pm$ 0.023} & 0.620 $\pm$ 0.003 & 0.502 $\pm$ 0.003 & \cellcolor{red!15}\textbf{\textcolor{blue}{0.812 $\pm$ 0.012}} \\
    &GIN & \cellcolor{green!15}0.867 $\pm$ 0.005 & 0.856 $\pm$ 0.006 & 0.839 $\pm$ 0.004 & \cellcolor{red!15}\textbf{0.858 $\pm$ 0.005} & 0.655 $\pm$ 0.024 & 0.691 $\pm$ 0.011 & \cellcolor{red!15}\textbf{0.789 $\pm$ 0.013} & 0.857 $\pm$ 0.005 & 0.844 $\pm$ 0.005 & \cellcolor{red!15}\textbf{\textcolor{blue}{0.891 $\pm$ 0.002}} \\
    \rowcolor{mygray}
    \cellcolor{white}
    \multirow{-4}{*}{\rotatebox[origin=c]{90}{\textbf{CS}}}
    &SAGE & \cellcolor{green!15}0.932 $\pm$ 0.004 & 0.896 $\pm$ 0.005 & 0.896 $\pm$ 0.003 & \cellcolor{red!15}\textbf{0.905 $\pm$ 0.004} & 0.745 $\pm$ 0.009 & 0.679 $\pm$ 0.003 & \cellcolor{red!15}\textbf{0.886 $\pm$ 0.010} & 0.904 $\pm$ 0.007 & 0.903 $\pm$ 0.001 & \cellcolor{red!15}\textbf{\textcolor{blue}{0.927 $\pm$ 0.002}} \\
    \midrule
    &GAT & \cellcolor{green!15}0.955 $\pm$ 0.005 & 0.917 $\pm$ 0.001 & 0.915 $\pm$ 0.001 & \cellcolor{red!15}\textbf{0.920 $\pm$ 0.002} & 0.871 $\pm$ 0.032 & 0.858 $\pm$ 0.044 & \cellcolor{red!15}\textbf{0.921 $\pm$ 0.004} & 0.920 $\pm$ 0.001 & 0.917 $\pm$ 0.000 & \cellcolor{red!15}\textbf{\textcolor{blue}{0.925 $\pm$ 0.001}} \\
    \rowcolor{mygray}
    \cellcolor{white}
    &GCN & \cellcolor{green!15}0.947 $\pm$ 0.002 & 0.597 $\pm$ 0.001 & 0.533 $\pm$ 0.001 & \cellcolor{red!15}\textbf{0.747 $\pm$ 0.010} & 0.817 $\pm$ 0.003 & 0.770 $\pm$ 0.001 & \cellcolor{red!15}\textbf{\textcolor{blue}{0.858 $\pm$ 0.008}} & 0.575 $\pm$ 0.003 & 0.506 $\pm$ 0.001 & \cellcolor{red!15}\textbf{0.815 $\pm$ 0.001} \\
    &GIN & \cellcolor{green!15}0.934 $\pm$ 0.003 & 0.903 $\pm$ 0.002 & 0.916 $\pm$ 0.001 & \cellcolor{red!15}\textbf{0.921 $\pm$ 0.002} & 0.842 $\pm$ 0.009 & 0.840 $\pm$ 0.006 & \cellcolor{red!15}\textbf{0.907 $\pm$ 0.003} & 0.924 $\pm$ 0.002 & 0.919 $\pm$ 0.001 & \cellcolor{red!15}\textbf{\textcolor{blue}{0.926 $\pm$ 0.001}} \\
    \rowcolor{mygray}
    \cellcolor{white}
    \multirow{-4}{*}{\rotatebox[origin=c]{90}{\textbf{Physics}}}
    &SAGE & \cellcolor{green!15}0.956 $\pm$ 0.005 & 0.712 $\pm$ 0.003 & 0.717 $\pm$ 0.002 & \cellcolor{red!15}\textbf{0.823 $\pm$ 0.011} & 0.905 $\pm$ 0.003 & 0.894 $\pm$ 0.003 & \cellcolor{red!15}\textbf{0.922 $\pm$ 0.001} & 0.926 $\pm$ 0.003 & 0.924 $\pm$ 0.002 & \cellcolor{red!15}\textbf{\textcolor{blue}{0.933 $\pm$ 0.001}} \\
    \bottomrule
    \end{tabular}
}
\end{table*}

\subsection{Evaluation of Model Utility}
\label{subsection:model_utility_evaluation}
Next, we evaluate the model utility of different graph unlearning methods.
\autoref{table:node_unlearning_utility} (the red ground columns) shows the experimental results for node unlearning.
For a fair comparison, we also apply \optaggr for \random.

\begin{table}[!t]
    \centering
    \caption{
    Comparison of F1 scores for MLP and four GNN models.
    A larger gap in F1 scores for MLP and GNN models means that the graph structural information is more important for the GNN models.
    }
    \label{table:mlp}
    \footnotesize
    \resizebox{0.95\linewidth}{!}{
    \setlength{\tabcolsep}{0.3em}
    \begin{tabular}{c | c c c c c}
    \toprule
    \textbf{Model} & \textbf{Cora} & \textbf{Citeseer} & \textbf{Pubmed} & \textbf{CS} & \textbf{Physics}\\
    \toprule
    \textbf{MLP} & 0.657 $\pm$ 0.019 & 0.587 $\pm$ 0.029 & 0.868 $\pm$ 0.002 & 0.927 $\pm$ 0.007 & 0.950 $\pm$ 0.003 \\
    \midrule
    \rowcolor{mygray}
    \textbf{GAT} & 0.823 $\pm$ 0.006 & 0.691 $\pm$ 0.015 & 0.851 $\pm$ 0.004 & 0.919 $\pm$ 0.004 & 0.955 $\pm$ 0.005 \\
    \textbf{GCN} & 0.739 $\pm$ 0.003 & 0.493 $\pm$ 0.006 & 0.748 $\pm$ 0.017 & 0.903 $\pm$ 0.006 & 0.947 $\pm$ 0.002 \\
    \rowcolor{mygray}
    \textbf{GIN} & 0.787 $\pm$ 0.013 & 0.587 $\pm$ 0.031 & 0.837 $\pm$ 0.015 & 0.867 $\pm$ 0.005 & 0.934 $\pm$ 0.003 \\
    \textbf{SAGE} & 0.824 $\pm$ 0.004 & 0.668 $\pm$ 0.013 & 0.874 $\pm$ 0.003 & 0.932 $\pm$ 0.004 & 0.956 $\pm$ 0.005 \\
    \bottomrule
    \end{tabular}
    }
\end{table}

\mypara{Influence of Datasets}
We first observe that on the Cora and Citeseer datasets, our proposed method, \sysname-\bkm and \sysname-\blpa, can achieve a much better F1 score compared to the \random method.
For instance, on the \gcn model trained on the Cora dataset, the F1 score for \sysname-\blpa is $0.676$, while the corresponding result is $0.509$ for \random.
For the Pubmed, CS, and Physics datasets, the F1 score of the \random method is comparable to \sysname-\bkm and \sysname-\blpa, and can even achieve a similar F1 score as the \scratch method in some settings.
We conjecture this is due to the different contributions of the graph structural information to the utility of GNN models.
Intuitively, if the graph structural information does not contribute much to the GNN models, it is not surprising that the \random method can achieve comparable model utility as \sysname-\blpa and \sysname-\bkm.

To validate whether the graph structural information indeed diversely affects the GNN models' performance among different datasets, we introduce a baseline that uses a 3-layer MLP (multi-layer perceptron) to train the prediction models for all datasets.
Note that we only use the node features to train the MLP model, without considering any graph structural information.
\autoref{table:mlp} depicts the comparison of the F1 scores between the MLP model and four GNN models on five datasets.
We observe that for the Cora and Citeseer datasets, the F1 score of the MLP model is significantly lower than that of the GNN models, which means the graph structural information plays a major role in the GNN models.
On the other hand, the MLP model can achieve adequate F1 score compared to the GNN models on Pubmed, CS, and Physics datasets, which means the graph structural information does not contribute much in the GNN models.

In conclusion, the contribution of the graph structural information to the GNN model can significantly affect the behaviors of different shard-based graph unlearning methods.

\mypara{Guideline for Choosing an Unlearning Method}
In practice, we would suggest the model provider evaluate the role of graph structure before choosing a proper graph unlearning method.
To this end, they can first compare the F1 score of MLP and GNN, if the gap in the F1 score between MLP and GNN is small, the \random method can be a good choice since it is much easier to implement, and it can achieve comparable model utility as \sysname-\blpa and \sysname-\bkm.
Otherwise, \sysname-\blpa and \sysname-\bkm are better choices due to better model utility.

Regarding the choice between the two shard partition methods, i.e., \sysname-\blpa and \sysname-\bkm, we empirically observe that if the GNN follows the \gcn structure, one can choose \sysname-\blpa, otherwise, one can adopt \sysname-\bkm.
We posit this is because the \gcn model requires the node degree information for normalization (see \autoref{subsection:gnn}), and the \sysname-\blpa can preserve more local structural information thus better preserve the node degree~\cite{WL20a}.

\mypara{Comparison with \scratch}
Interestingly, we could observe that \sysname-\bkm performs slightly better than \scratch in some cases.
For instance, the F1 score of \sysname-\bkm is 0.801 on the Cora dataset and the \gin model, while the corresponding F1 score of \scratch is 0.787.
There are two possible reasons for this phenomenon. 
First, sampling often can eliminate some ``noise'' in the dataset, which is consistent with the observation of prior studies~\cite{ZZSKP20,CLSLBH19}. 
Second, \sysname makes the final prediction by aggregating all submodels' results, in this sense, \sysname performs an ensemble, another way to improve model performance.

\smallskip
Considering the conclusions for node unlearning and edge unlearning are similar in terms of both unlearning efficiency and model utility, the results can be found at \autoref{app:edge_unlearning_results}, we only provide the results for node unlearning in the following parts.

\subsection{Effectiveness of \optaggr}
\label{subsection:optaggr_effectiveness}
To validate the effectiveness of the \optaggr method proposed in \autoref{section:optimal_aggregation}, we compare with \meanaggr and \majaggr by conducting experiments on five datasets and four GNN models.
\autoref{table:node_unlearning_utility} illustrates the F1 scores of different aggregation methods for \scratch, \sysname-\blpa, and \sysname-\bkm.

\mypara{Observations}
In general, \optaggr can effectively improve the F1 score in most cases compared to \meanaggr and \majaggr.
For instance, on the Cora dataset with \sysname-\blpa unlearning method, \optaggr achieves $0.357$ higher F1 score than that of \majaggr for the \gcn model.
We also observe that the \majaggr method performs the worst in most cases.
We posit it is because \majaggr neglects information of the posteriors obtained from each shard model.
Concretely, if the posteriors of the shard models have high confidence to multiple classes rather than a single class, the \majaggr method will lose information about the runner-up classes, leading to bad model utility.

Comparing different GNN models, \gcn benefits the most while \gin benefits the least from \optaggr.
In terms of model utility, the \sysname-\blpa method benefits the most from \optaggr.
We conjecture this is because the \blpa partition method can capture the local structural information while losing some of the global structural information of the training graph~\cite{SHFZWS20a,WL20a}.
Using \optaggr helps better capture the global structural information by assigning different importance scores to shard models.

\begin{table}[!tbp]
    \centering
    \caption{Impact of the number of training nodes for learning \optaggr.
    ``10\%'' and ``1000'' stand for randomly selecting 10\% and 1000 nodes from the training graph, respectively. ``All'' stands for using all nodes in the training graph.
    We highlight our recommended choices in the \colorbox{red!15}{red} ground.
    }
    \label{table:opt_samples}
    \resizebox{1\linewidth}{!}{
    \footnotesize
    \setlength{\tabcolsep}{0.4em}
    \begin{tabular}{c c | c c | c c c}
    \toprule
    \textbf{$\model$} & \#. \textbf{Nodes} & \textbf{Cora} & \textbf{Citeseer} & \textbf{Pubmed} & \textbf{CS} & \textbf{Physics}\\
    \toprule
	 & \textbf{10\%} & \cellcolor{red!15}0.70 $\pm$ 0.02 & \cellcolor{red!15}0.71 $\pm$ 0.01 & 0.86 $\pm$ 0.00 & 0.91 $\pm$ 0.00 & 0.93 $\pm$ 0.00 \\
	 & \textbf{1000} & 0.73 $\pm$ 0.01 & 0.72 $\pm$ 0.02 & \cellcolor{red!15}0.86 $\pm$ 0.00 & \cellcolor{red!15}0.91 $\pm$ 0.01 & \cellcolor{red!15}0.93 $\pm$ 0.00 \\
	\multirow{-3}{*}{\rotatebox[origin=c]{90}{\textbf{GAT}}}
	 & \textbf{All} & 0.74 $\pm$ 0.00 & 0.72 $\pm$ 0.00 & 0.86 $\pm$ 0.00 & 0.91 $\pm$ 0.00 & 0.93 $\pm$ 0.00 \\
	\midrule
	 & \textbf{10\%}  & \cellcolor{red!15}0.44 $\pm$ 0.00 & \cellcolor{red!15}0.31 $\pm$ 0.01 & 0.48 $\pm$ 0.00 & 0.81 $\pm$ 0.00 & 0.82 $\pm$ 0.00 \\
	 & \textbf{1000} & 0.49 $\pm$ 0.01 & 0.31 $\pm$ 0.02 & \cellcolor{red!15}0.47 $\pm$ 0.01 & \cellcolor{red!15}0.81 $\pm$ 0.00 & \cellcolor{red!15}0.80 $\pm$ 0.00 \\
	\multirow{-3}{*}{\rotatebox[origin=c]{90}{\textbf{GCN}}}
	 & \textbf{All} & 0.50 $\pm$ 0.00 & 0.32 $\pm$ 0.03 & 0.48 $\pm$ 0.00 & 0.82 $\pm$ 0.01 & 0.81 $\pm$ 0.01 \\
	\midrule
	 & \textbf{10\%}  & \cellcolor{red!15}0.70 $\pm$ 0.00 & \cellcolor{red!15}0.72 $\pm$ 0.00 & 0.86 $\pm$ 0.00 & 0.88 $\pm$ 0.00 & 0.93 $\pm$ 0.00 \\
	 & \textbf{1000} & 0.72 $\pm$ 0.02 & 0.73 $\pm$ 0.02 & \cellcolor{red!15}0.86 $\pm$ 0.00 & \cellcolor{red!15}0.89 $\pm$ 0.00 & \cellcolor{red!15}0.91 $\pm$ 0.03 \\
	\multirow{-3}{*}{\rotatebox[origin=c]{90}{\textbf{GIN}}}
	 & \textbf{All} & 0.76 $\pm$ 0.00 & 0.71 $\pm$ 0.00 & 0.86 $\pm$ 0.00 & 0.89 $\pm$ 0.00 & 0.93 $\pm$ 0.00 \\
	\midrule
	 & \textbf{10\%}  & \cellcolor{red!15}0.71 $\pm$ 0.01 & \cellcolor{red!15}0.70 $\pm$ 0.00 & 0.87 $\pm$ 0.00 & 0.93 $\pm$ 0.00 & 0.94 $\pm$ 0.00 \\
	 & \textbf{1000} & 0.73 $\pm$ 0.03 & 0.71 $\pm$ 0.00 & \cellcolor{red!15}0.87 $\pm$ 0.00 & \cellcolor{red!15}0.92 $\pm$ 0.00 & \cellcolor{red!15}0.93 $\pm$ 0.01 \\
	\multirow{-3}{*}{\rotatebox[origin=c]{90}{\textbf{SAGE}}}
	 & \textbf{All} & 0.74 $\pm$ 0.00 & 0.72 $\pm$ 0.00 & 0.87 $\pm$ 0.00 & 0.92 $\pm$ 0.00 & 0.94 $\pm$ 0.00 \\
    \bottomrule
    \end{tabular}
}
\end{table}

\mypara{Impact of the Number of Training Nodes}
As discussed in \autoref{section:optimal_aggregation}, to further improve the unlearning efficiency, one can use a small portion of nodes in the training graph to learn the importance score.
Doing this can effectively reduce the relearning time of \optaggr, as shown in \autoref{subsection:unlearning_efficiency_evaluation}.
Here we evaluate its impact on the model utility.
We experiment on three different cases: randomly sample $10\%$ of nodes, randomly sample a fixed number of 1,000 nodes, and use all nodes, in the training graph.

\autoref{table:opt_samples} illustrates the results on five datasets and four GNN models for \sysname-\bkm.
We observe that both using 10\% of nodes and using a fixed number of 1,000 nodes can achieve comparable model utility as that of using all nodes.
In practice, we suggest the model provider to adopt the minimum of 10\% and 1,000 to learn the importance scores.
In another word, the model provider can use 10\% for small graphs, and use 1,000 for large graphs.
The conclusions are the same for \sysname-\blpa.

\begin{table*}[!t]
    \centering
    \caption{
    Comparison of F1 scores for different graph partition methods.
    We highlight our proposed method in the \colorbox{red!15}{red} ground and the best results in bold.}
    \label{table:baseline_utility}
    \footnotesize
    \setlength{\tabcolsep}{1.2em}
    \begin{tabular}{cc|cc|cc|c}
    \toprule
    \textbf{Dataset} & \textbf{Model} & \multicolumn{2}{c|}{\textbf{BLPA-based}} & \multicolumn{2}{c|}{\textbf{BEKM-based}} & \multicolumn{1}{c}{\textbf{Minimum Edge Cut}}\\
   \dset & \model & \sysname-\blpa & \blpa-LP & \sysname-\bkm  & \bkm-Hungarian & METIS \\
    \toprule
	&\textbf{GAT}& \cellcolor{red!15}0.727 $\pm$ 0.009 & 0.712 $\pm$ 0.006 & \cellcolor{red!15}\textbf{{0.754 $\pm$ 0.009}} & 0.740 $\pm$ 0.006 & 0.683 $\pm$ 0.007 \\
	&\textbf{GCN}& \cellcolor{red!15}\textbf{{0.676 $\pm$ 0.004}} & 0.668 $\pm$ 0.020 & \cellcolor{red!15}0.531 $\pm$ 0.009 & 0.552 $\pm$ 0.005 & 0.458 $\pm$ 0.010 \\
	&\textbf{GIN}& \cellcolor{red!15}0.753 $\pm$ 0.015 & 0.722 $\pm$ 0.029 & \cellcolor{red!15}\textbf{{0.801 $\pm$ 0.018}} & 0.795 $\pm$ 0.016 & 0.703 $\pm$ 0.020 \\
	\multirow{-4}{*}{\rotatebox[origin=c]{90}{\textbf{Cora}}}
	&\textbf{SAGE}& \cellcolor{red!15}0.684 $\pm$ 0.014 & 0.708 $\pm$ 0.002 & \cellcolor{red!15}\textbf{{0.740 $\pm$ 0.013}} & 0.739 $\pm$ 0.005 & 0.694 $\pm$ 0.008 \\
	\midrule
	&\textbf{GAT}& \cellcolor{red!15}0.688 $\pm$ 0.005 & 0.590 $\pm$ 0.009 & \cellcolor{red!15}\textbf{{0.738 $\pm$ 0.006}} & 0.737 $\pm$ 0.003 & 0.615 $\pm$ 0.002 \\
	&\textbf{GCN}& \cellcolor{red!15}\textbf{{0.516 $\pm$ 0.004}} & 0.504 $\pm$ 0.022 & \cellcolor{red!15}0.417 $\pm$ 0.018 & 0.397 $\pm$ 0.023 & 0.457 $\pm$ 0.006 \\
	&\textbf{GIN}& \cellcolor{red!15}0.597 $\pm$ 0.021 & 0.589 $\pm$ 0.041 & \cellcolor{red!15}\textbf{{0.678 $\pm$ 0.072}} & 0.655 $\pm$ 0.059 & 0.574 $\pm$ 0.064 \\
	\multirow{-4}{*}{\rotatebox[origin=c]{90}{\textbf{Citeseer}}}
	&\textbf{SAGE}& \cellcolor{red!15}0.642 $\pm$ 0.005 & 0.682 $\pm$ 0.007 & \cellcolor{red!15}\textbf{{0.743 $\pm$ 0.002}} & 0.734 $\pm$ 0.002 & 0.677 $\pm$ 0.004 \\
	\midrule
	&\textbf{GAT}& \cellcolor{red!15}0.858 $\pm$ 0.003 & 0.857 $\pm$ 0.001 & \cellcolor{red!15}\textbf{{0.860 $\pm$ 0.003}} & 0.857 $\pm$ 0.003 & 0.841 $\pm$ 0.001  \\
	&\textbf{GCN}& \cellcolor{red!15}\textbf{{0.718 $\pm$ 0.010}} & 0.709 $\pm$ 0.004 & \cellcolor{red!15}0.659 $\pm$ 0.020 & 0.628 $\pm$ 0.034 & 0.650 $\pm$ 0.018 \\
	&\textbf{GIN}& \cellcolor{red!15}0.855 $\pm$ 0.004 & 0.854 $\pm$ 0.001 & \cellcolor{red!15}\textbf{{0.859 $\pm$ 0.003}} & 0.853 $\pm$ 0.001 & 0.836 $\pm$ 0.001 \\
	\multirow{-4}{*}{\rotatebox[origin=c]{90}{\textbf{Pubmed}}}
	&\textbf{SAGE}& \cellcolor{red!15}0.863 $\pm$ 0.002 & 0.857 $\pm$ 0.003 & \cellcolor{red!15}\textbf{{0.862 $\pm$ 0.002}} & 0.858 $\pm$ 0.00 & 0.849 $\pm$ 0.003 \\
	\midrule
	&\textbf{GAT}& \cellcolor{red!15}0.858 $\pm$ 0.004 & 0.862 $\pm$ 0.003 & \cellcolor{red!15}\textbf{{0.906 $\pm$ 0.002}} & 0.901 $\pm$ 0.003 & 0.891 $\pm$ 0.013 \\
	&\textbf{GCN}& \cellcolor{red!15}\cellcolor{red!15}0.750 $\pm$ 0.023 & 0.745 $\pm$ 0.004 & \cellcolor{red!15}\textbf{{0.812 $\pm$ 0.012}} & 0.806 $\pm$ 0.007 & 0.782 $\pm$ 0.021 \\
	&\textbf{GIN}& \cellcolor{red!15}0.789 $\pm$ 0.013 & 0.786 $\pm$ 0.003 & \cellcolor{red!15}\textbf{{0.891 $\pm$ 0.002}}& 0.883 $\pm$ 0.007 & 0.862 $\pm$ 0.002 \\
	\multirow{-4}{*}{\rotatebox[origin=c]{90}{\textbf{CS}}}
	&\textbf{SAGE}& \cellcolor{red!15}0.886 $\pm$ 0.010 & 0.889 $\pm$ 0.023 & \cellcolor{red!15}\textbf{{0.927 $\pm$ 0.002}}& 0.922 $\pm$ 0.002 & 0.906 $\pm$ 0.004 \\
	\midrule
	&\textbf{GAT}& \cellcolor{red!15}0.921 $\pm$ 0.004 & 0.918 $\pm$ 0.004 & \cellcolor{red!15}\textbf{{0.925 $\pm$ 0.001}} & 0.923 $\pm$ 0.001 & 0.918 $\pm$ 0.002 \\
	&\textbf{GCN}& \cellcolor{red!15}\textbf{{0.858 $\pm$ 0.008}} & 0.856 $\pm$ 0.005 & \cellcolor{red!15}0.815 $\pm$ 0.001 & 0.808 $\pm$ 0.001 & 0.810 $\pm$ 0.001 \\
	&\textbf{GIN}& \cellcolor{red!15}0.907 $\pm$ 0.003 & 0.897 $\pm$ 0.011 & \cellcolor{red!15}\textbf{{0.926 $\pm$ 0.001}} & 0.923 $\pm$ 0.002 & 0.895 $\pm$ 0.003 \\
	\multirow{-4}{*}{\rotatebox[origin=c]{90}{\textbf{Physics}}}
	&\textbf{SAGE}& \cellcolor{red!15}0.922 $\pm$ 0.001 & 0.913 $\pm$ 0.002 & \cellcolor{red!15}\textbf{{0.933 $\pm$ 0.001}} & 0.931 $\pm$ 0.001 & 0.911 $\pm$ 0.005 \\
    \bottomrule
\end{tabular}
\end{table*}

\begin{table}[!t]
    \centering
    \caption{
    Comparison of graph partition efficiency for different balanced graph partition methods.
    We highlight our proposed partition methods in the \colorbox{red!15}{red} ground.}
    \label{table:baseline_efficiency}
    \resizebox{1\linewidth}{!}{
    \setlength{\tabcolsep}{0.2em}
    \begin{tabular}{c|cc|cc|c}
        \toprule
        \textbf{Dataset} & \multicolumn{2}{c|}{\textbf{BLPA-based}} & \multicolumn{2}{c|}{\textbf{BEKM-based}} & \multicolumn{1}{c}{\textbf{Minimum Edge Cut}}\\
       \dset & \sysname & LP & \sysname  & Hungarian & METIS \\        
        \toprule
        \textbf{Cora} & \cellcolor{red!15}\textbf{{3s}} & 179s & \cellcolor{red!15}\textbf{{26s}} & 817s & 4s \\
        \textbf{Citeseer} & \cellcolor{red!15}\textbf{{2s}} & 30s & \cellcolor{red!15}\textbf{{20s}} & 1,309s & 3s \\
        \textbf{Pubmed} & \cellcolor{red!15}\textbf{{20s}} & 301s & \cellcolor{red!15}\textbf{{240s}} & 174,684s & 21s \\
        \textbf{CS} & \cellcolor{red!15}\textbf{{13s}} & 705s & \cellcolor{red!15}\textbf{{220s}} & 174,498s & 15s \\
        \textbf{Physics} & \cellcolor{red!15}\textbf{{40s}} & 2,351s & \cellcolor{red!15}\textbf{{480s}} & 948,790s & 58s \\
        \bottomrule
    \end{tabular}
}
\end{table}

\subsection{Comparison with Existing Balanced Graph Partition Solutions}
\label{subsection:comparison_balanced_partition}
In this section, we empirically compare \sysname with existing solutions for balanced graph partitioning~\cite{UB13,LK16,MF14} in terms of running time and model utility.

\mypara{Competitors}
These algorithms can be broadly classified into three categories:
The first category considers only the graph structural information and relies on community detection as \sysname-\blpa.
The second category considers the graph structural information without relying on community detection.
The third category considers both structural information and node features as \sysname-\bkm.
For each category, we choose one most representative method as competitor, and we list the details as follows.
\begin{itemize}[leftmargin=*]
    \item \textbf{BLPA-LP~\cite{UB13}.}
    Similar to our proposed \sysname-\blpa, this method achieves a balanced graph partition by constraining the label propagation process.
    The general idea is to formulate the label propagation process as a linear programming problem with $2k(k-1)$ variables and $2k^2 + nk(k - 1)$ constraints, where $n$ and $k$ are the number of nodes and the number of shards, respectively.
    When the size of the graph and the number of shards are large, solving the linear programming problem is time-consuming.
    \item \textbf{METIS~\cite{LK16}.}
    The objective of METIS is to obtain the balanced graph partition while cutting the minimum number of edges.
    The computational complexity of METIS is $\mathcal{O}((n+m)\cdot \log{k})$, where $m$ is the number of edges. 
    We implement this method with official METIS 5.1.0\footnote{\url{http://glaros.dtc.umn.edu/gkhome/metis/metis/overview}} and a Python wrapper\footnote{\url{https://github.com/inducer/pymetis}} for METIS library.
    \item \textbf{BEKM-Hungarian~\cite{MF14}.}
    BEKM-Hungarian shares the general idea of our \sysname-\bkm.
    The main difference is that it has a different mechanism in the node reassignment step for achieving balanced $k$-means.
    Concretely, BEKM-Hungarian formulates the node reassignment step as a matching problem and is approximately solved by the Hungarian algorithm.
    The computational complexity of the Hungarian algorithm is $\mathcal{O}(n^3)$.
\end{itemize}

\noindent The reason why we choose these three algorithms is that they achieve the-state-of-the-art performance for each category.
We discuss other existing algorithms and how these three algorithms fit into the whole balanced graph partitioning field in \autoref{section:related}.

\mypara{Results}
\autoref{table:baseline_utility} and \autoref{table:baseline_efficiency} illustrate the model utility and graph partitioning efficiency for different methods.
We apply \optaggr for all the graph partitioning methods for a fair comparison.

In general, graph partitioning methods rely on both graph structural information and node features. i.e., \sysname-\bkm and BEKM-Hungarian, achieve the best model utility when the target model is GAT, GIN, and SAGE, which is consistent with the conclusion of \autoref{subsection:model_utility_evaluation}.
Comparing \sysname-\bkm and BEKM-Hungarian, we observe that they achieve similar model utility; however, the computational complexity of BEKM-Hungarian ($\mathcal{O}(n^3)$) is much higher than that of \sysname-\bkm ($\mathcal{O}(k\cdot n)$).
From \autoref{table:baseline_efficiency}, we also observe that BEKM-Hungarian is not scalable to large graphs.

When the target model is GCN, the community detection-based methods, i.e., \sysname-\blpa and BLPA-LP, achieve a better model utility than the minimum-cut-based method (METIS).
We suspect this is because the \gcn model requires the node degree information for normalization, and the community detection-based methods can preserve more local structural information, thus better preserving the node degree.
Comparing \sysname-\blpa and BLPA-LP, \sysname-\blpa is more efficient than BLPA-LP (see \autoref{table:baseline_efficiency}) while achieving comparable model utility.

\mypara{Remarks}
\sysname is a general framework for GNN unlearning; any balanced graph partitioning method which meets the requirements in \autoref{subsection:problem_definition} can be adopted.
Therefore, we encourage the community to develop more efficient and better performing balanced graph partitioning algorithms for graph unlearning.

\begin{table}
    \centering
    \caption{Attack AUC of membership inference against our \sysname ($\attack_{I}$) and deterministic unlearning ($\attack_{II}$).}
    \label{table:attack_unlearning}
    \vspace{-0.1cm}
    \footnotesize
    \resizebox{1\linewidth}{!}{
    \setlength{\tabcolsep}{0.7em}
    \begin{tabular}{c | c c | c c | c c | c c}
    \toprule
    \multicolumn{1}{c|}{\textbf{Model}}
    & \multicolumn{2}{c|}{\textbf{GAT}}
    & \multicolumn{2}{c|}{\textbf{GCN}} 
    & \multicolumn{2}{c|}{\textbf{GIN}}
    & \multicolumn{2}{c}{\textbf{SAGE}} \\
    \multicolumn{1}{c|}{\textbf{Dataset}} 
    & $\attack_{I}$ & $\attack_{II}$ & $\attack_{I}$ & $\attack_{II}$ & $\attack_{I}$ & $\attack_{II}$ & $\attack_{I}$ & $\attack_{II}$ \\
    \toprule
    \multirow{1}{*}{\rotatebox[origin=c]{0}{\textbf{Cora}}}
    & 0.512 & 0.508 & 0.511 & 0.510 & 0.513 & 0.510 & 0.511 & 0.510 \\
    \rowcolor{mygray}
    \multirow{1}{*}{\rotatebox[origin=c]{0}{\textbf{Citeseer}}}
    & 0.515 & 0.510 & 0.510 & 0.510 & 0.513 & 0.513 & 0.512 & 0.510 \\
    \multirow{1}{*}{\rotatebox[origin=c]{0}{\textbf{Pubmed}}}
    & 0.509 & 0.510 & 0.511 & 0.509 & 0.512 & 0.511 & 0.510 & 0.511 \\
    \rowcolor{mygray}
    \multirow{1}{*}{\rotatebox[origin=c]{0}{\textbf{CS}}}
    & 0.510 & 0.509 & 0.520 & 0.511 & 0.515 & 0.514 & 0.515 & 0.513 \\
    \multirow{1}{*}{\rotatebox[origin=c]{0}{\textbf{Physics}}}
    & 0.519 & 0.515 & 0.518 & 0.512 & 0.512 & 0.510 & 0.517 & 0.517 \\
    \bottomrule
    \end{tabular}
    }
\end{table}

\subsection{Unlearning Power of \sysname}
\label{subsec:unlearning_power}
Since our method is highly empirical, we adopt the state-of-the-art attack against machine unlearning~\cite{CZWBHZ21} to quantify the extra information leakage of \sysname when the graph is not re-partitioned.
In particular, Chen et al.~\cite{CZWBHZ21} showed that the attackers, using an enhanced membership inference attack~\cite{SSSS17}, can determine whether a target sample exists in the original model and is revoked from the unlearned model when they have access to both the original and unlearned model.
Here we quantify the {extra information leakage} of \sysname as {\it the attack's performance difference between deterministic unlearning and \sysname unlearning}.
Concretely, we introduce two scenarios of membership inference attacks.  
We start from the same set of original shard models.  
In scenario 1, the unlearned models are obtained by directly deleting the revoked nodes from the corresponding shard graph, and retrain the corresponding shard models.
This is how \sysname generates the unlearned models.
In scenario 2, we retrain from scratch (re-partition the graph, and train a set of new shard models).
This type of unlearning deterministically unlearn every component while it is extremely time-consuming.
Denoting the two scenarios as $\attack_{I}$ and $\attack_{II}$, the extra information leakage is the difference of the attack AUC between $\attack_{I}$ and $\attack_{II}$. 
We use the implementation\footnote{\url{https://github.com/MinChen00/UnlearningLeaks}} of~\cite{CZWBHZ21} to conduct our experiments.
The experimental results in \autoref{table:attack_unlearning} show that the attack AUC of both $\attack_{I}$ and $\attack_{II}$ are close to 0.5 (random guessing), meaning that \sysname does not leak much extra information.
This is also consistent with the observation of~\cite{CZWBHZ21} that the membership inference performs bad on \sisa based method due to the fact that the aggregation reduces the influence of a specific sample on its global model.

\section{Discussion}
\label{section:discussion}

\mypara{Guarantee to the right to be forgotten}
Note that when the adversaries have access to both the original and the unlearned models, the presence of the deleted node might be inferred using the friendship information of the graph.
A previous study~\cite{CZWBHZ21} has shown that machine unlearning is vulnerable to membership inference attack.
However, these attacks are orthogonal to our work since the primary goal of machine unlearning is to comply with ``legitimate regulations'' such as the GDPR. 
In this sense, as long as the model is trained without the revoked sample, the requirement of the right to be forgotten is satisfied.
To mitigate the potential attacks, one can deploy some defense mechanisms as discussed in~\cite{CZWBHZ21}, which can be add-ons of \sysname.

\mypara{Compatibility with Commercial Graph Services}
Compared with the existing graph-learning-based services, the additional cost of \sysname is graph partition; however, once the partition is defined, we can keep it fixed without extra effort. 
The process of training shard models is the same as the existing services. 
Once this pipeline is built, the maintenance effort of dealing with unlearning requests is much lower than existing services, since \sysname only needs to retrain the sub-model containing the deleted samples.

\mypara{Additional Cost of Maintaining Massive Shards}
One might argue that maintaining the shard models is more expensive than maintaining one global model. 
However, in machine unlearning, the cost of retraining the global model is much higher than maintaining the shard models.
Comparing to maintaining one global model, the additional cost of maintaining the shard models comes from two sources: 
(1) Additional prediction time cost due to the aggregation process;
(2) Additional storage cost of storing multiple shard models instead of storing one global model.
From the time cost perspective, we have empirically shown in \autoref{subsection:unlearning_efficiency_evaluation} that the additional prediction time introduced by the shard-based methods is much less than the retraining time of the global model.
From the economic cost perspective, it is well-known that the computation cost (of retraining the global model) is much higher than the storage cost (of renting disk for storing the shard models)~\cite{BCCJTZLP21}.
For instance, the storage costs are of $\$0.026/GB$ per month on
Google Cloud, $\$0.023/GB$ per month on Amazon Web Services,
and $\$0.018/GB$ per month on Azure at the time of writing.
Instead, renting the cheapest GPUs starts at $\$0.35/hour$ on
Google Cloud, $\$0.526/hour$ on Amazon Web Services, and
$\$0.90/hour$ on Azure.

\mypara{Adaptive Machine Unlearning}
Gupta et. al~\cite{GJNRSW21} define the notion of $(\alpha,\beta,\gamma)$-unlearning, which enforces that the output of any unlearning algorithm should be similar to retraining from scratch.
The authors prove that the general family of distributed learning and unlearning algorithms such as \sisa method satisfies $(\alpha,\beta,\gamma)$-unlearning in the {\it non-adaptive} setting (unlearning requests arrive in a non-adaptive way), but it does not satisfy $(\alpha,\beta,\gamma)$-unlearning in the {\it adaptive} setting.
As \sysname \xspace belongs to this general family, \sysname also satisfies $(\alpha,\beta,\gamma)$-unlearning in the non-adaptive setting, but does not satisfy the adaptive setting.

\mypara{Handling Different Scenarios}
In real-world applications, nodes are also likely to be inserted into the training graph of the GNN model. 
We can insert the node to the shard containing the highest number of its neighbors and retrain the corresponding shard model.
We refer the readers to \autoref{app:community_depedent} for handling a scenario where the removal request comes independently from a specific community.

\section{Related Work}
\label{section:related}

\mypara{Machine Unlearning}
The notion of machine unlearning was first proposed by Cao et al.~\cite{CY15}.
Subsequently, the research in machine unlearning has proceeded into two directions: Deterministic unlearning and approximate unlearning.
The objective of \textit{deterministic unlearning} (some papers call it \textit{exact unlearning}) is to guarantee that the influence of the revoked samples are completely removed from the target model.
The straightforward approach of retraining the global model from scratch perfectly satisfies deterministic unlearning; however, it is computationally infeasible when the dataset is large.
To reduce the computational cost of retraining from scratch, Cao et al.~\cite{CY15} consider statistical query learning and dissect the model into a summation form, so that removing a sample can be done efficiently by subtracting the summand corresponds to that sample.
However, the algorithm in~\cite{CY15} only applies to learning algorithms that can be transformed into summation form, limiting itself not for neural networks.

Recently, Ginart et al.~\cite{GGVZ19} have proposed the notion of ($\epsilon$, $\delta$)-approximate unlearning in a way reminiscent of DP~\cite{DR14,ZWLHC18,ZWLHBHCZ21,WCZSCLLJ21,DZBLJCC21}.
It guarantees that the output distribution of the unlearned model is \textit{close} to the model trained without the revoked samples.
Formally, an unlearning algorithm $U_A$ satisfies ($\epsilon$, $\delta$)-approximate unlearning if $\Pr{A(D_{-i}) \in S | D_{-i}} \leq \epsilon \cdot \Pr{U_A(D, A(D), i) \in S | D_{-i}} + \delta$, where $A$ is the learning algorithm, $D$ is the training dataset, $S$ is the possible output of the model, and $i$ is the revoked sample.
In this sense, DP is a natural choice to support ($\epsilon$, $\delta$)-approximate unlearning.
However, when there are a group of samples to be deleted, we would need to use group DP, which greatly increases the amount of noise needed and decreases the model utility; thus, DP is not directly adopted to implement approximate unlearning~\cite{GGVZ19}.

While Ginart et al.~\cite{GGVZ19} proposed a ($\epsilon$, $\delta$)-approximate unlearning algorithm for the $k$-means problem, 
Guo et al.~\cite{GGHM20} gave approximate unlearning algorithms for linear and logistic regression.
It first performs a convex optimization step and is followed by a Gaussian perturbation.
The algorithm yields error that grows linearly with the number of updates.
Izzo et al.~\cite{ISCZ21} also focus on linear regression and show how to improve the run-time per deletion of~\cite{GGHM20} from quadratic to linear in the dimension.
Neel et al.~\cite{NRS21} leverages a distributed optimization that partitions the data, separately optimizes on each partition, and then averages the parameters.
It guarantees that, for a fixed accuracy target, the run-time of the update operation is constant in the length of the update sequence, and it can deal with all convex models.

Due to the strong theoretical requirement of ($\epsilon$, $\delta$)-approximate unlearning, most of previous studies can only deal with linear or convex models.
Note that GNN is a highly non-convex model, which makes it difficult to theoretically prove that \sysname satisfies ($\epsilon$, $\delta$)-approximate unlearning; thus, we empirically quantify the possible information leakage relying on membership inference.

\mypara{Balanced Graph Partitioning}
As discussed in \autoref{subsection:comparison_balanced_partition}, the existing balanced graph partitioning algorithms can be broadly classified into three categories.
The first two categories adopt \textbf{Strategy 1} in \autoref{section:graph_partition} that only consider graph structural information.
The third category adopt \textbf{Strategy 2} in \autoref{section:graph_partition} that consider both graph structural and node feature information.

Recall that in \autoref{subsection:community_detection}, community detection can inherently preserve graph structural information with the cost of unbalanced partitioning.
Thus, the first category of previous studies aim to modify existing community detection methods to satisfy balanced community size constraint.
In BLPA-LP~\cite{UB13}, the authors formulate the label propagation process as a linear programming problem to satisfy the community size constraints.

On the other hand, the second category of previous studies do not rely on community detection.
Instead, they directly partition the graph by optimizing some predefined criterion, such as minimizing the graph cut~\cite{SM97,DGRW11} or maximizing the graph modularity~\cite{N06,GMC10}.
However, these optimization problems are always NP-hard and cannot be solved exactly; thus, the researchers proposed many approximate or intuitive algorithms.
Spectral graph partitioning~\cite{M01,ZN15} is a widely adopted approach.
The general idea is first to calculate the Laplacian matrix of the graph, then calculate the eigenvectors of the Laplacian matrix.
Each node is mapped to one of the eigenvalues in the second smallest eigenvector, and the sign of the corresponding eigenvalues defines the graph partition.
One can conduct the spectral graph partitioning method hierarchically to partition the graph into multiple shards.
The main drawback of the spectral methods is they cannot deal with large-scale graphs.
A promising solution for large-scale graph partitioning is utilizing the multilevel graph partitioning methods.
The general idea is first to contract edges and obtain smaller graphs, then cut the resulting graph, and finally unfold back to the original graph with some local improvement criterion~\cite{ACL06,APY19,HLBL21}.
Among the multilevel graph partitioning methods, METIS~\cite{KK98,LK16} is a family of the most widely known techniques and achieves state-of-the-art performance~\cite{AU20}.

The general idea of the third category is first to transform the attributed graph into node embeddings and use balanced clustering methods to cluster the node embeddings.
In BEKM-Hungarian~\cite{MF14}, the authors modify the reassignment step of $k$-means algorithm to achieve balanced clusters.
The core idea is to formulate the node reassignment problem as a matching problem which is approximately solved by the Hungarian algorithm.
In~\cite{LHNL17}, the authors propose to use linear regression to estimate the class-specific hyperplanes that partition each class of the data point from others.
A soft balanced constraint is enforced to achieve balanced clustering.
The drawback of this method is that we cannot precisely control the cluster size.

\section{Conclusion}
\label{section:conclusion}
In this paper, we propose the first machine unlearning framework \sysname in the context of GNNs.
Concretely, we first identify two types of machine unlearning requests, namely node unlearning and edge unlearning.
We then propose a general pipeline for machine unlearning in GNN models.
To achieve efficient retraining while keeping the structural information of the graph, we propose a general principle for balancing the shards resulting from the graph partitioning and instantiate it with two novel balanced graph partition algorithms.
We further propose a learning-based aggregation method to improve the model utility.
Extensive experiments on five real-world graph datasets and four state-of-the-art GNN models illustrate the high unlearning efficiency and high model utility resulting from \sysname.

\section*{Acknowledgement}
\label{sec:ack}
We thank all anonymous reviewers for their constructive comments.
This work is partially funded by the Helmholtz Association within the project ``Trustworthy Federated Data Analytics'' (TFDA) (funding number ZT-I-OO1 4).
Tianhao Wang did part of the work while at Purdue University and Carnegie Mellon University, and was funded by National Science Foundation (NSF) grant No.1931443, a Bilsland Dissertation Fellowship, and a Packard Fellowship.

\bibliographystyle{abbrv}
\bibliography{clean.bib}
\smallskip
\appendix
\newpage

\section{Details of Graph Neural Networks}
\label{app:gnn_details}

\mypara{Message Passing}
We first initialize each node's embedding.
In the passing phase, every node receives a ``message'' from its neighbor nodes, and aggregates (there are different aggregation methods which we will talk about later) the messages as its intermediate embedding.
After $\ell$ steps, the embedding of a node can aggregate information from its $\ell$-hop neighbors.
Formally, during the $i$-th step, the embedding $\embed_u^{i}$ of node $u \in \nodeset$ is updated using information aggregated from $u$'s neighbors $\neigh{u}$ using a pair of \textit{aggregation operation} $\aggr$ and \textit{updating operation} $\upd$:
\begin{align*}
    \embed_u^{i+1} &= \update{i}{\embed_u^{i}, \messagee_{\neigh{u}}^i}\\
    \messagee_{\neigh{u}}^i&= \aggregate{i}{\embed_v^{i}, \forall v \in \neigh{u})}
\end{align*}
where $\messagee_{\neigh{u}}^i$ is the message received from $u$'s neighbors.

\mypara{Aggregation Operations}
We introduce four of the most widely used aggregation operations as follows:
\begin{itemize}[leftmargin=*]
    \item \mypara{Graph Isomorphism Networks (GIN)~\cite{XHLJ19}}
    \gin directly sums up the embeddings of $u$'s neighbors $\neigh{u}$, i.e., $\messagee_{\neigh{u}} = \sum_{v \in \neigh{u}} \embed_v$.

    \item \mypara{Graph SAmple and aggreGatE (SAGE)~\cite{HYL17}}
    \sage takes an average over $u$'s neighbors' embeddings rather than summing them up, i.e., $\messagee_{\neigh{u}} = \frac{\sum_{v \in \neigh{u}} \embed_v}{|\neigh{u}|}$.

    \item \mypara{Graph Convolution Networks (GCN)~\cite{KW17}}
    The \gcn method uses the symmetric normalization for aggregation, i.e.,
    $\messagee_{\neigh{u}} = \sum_{v \in \neigh{u}} \frac{\embed_v}{\sqrt{|\neigh{u}| \cdot |\neigh{v}|}}$.

    \item \mypara{Graph Attention Networks (GAT)~\cite{VCCRLB18}}
    \gat assigns an attention weight or importance score to each neighbor during the aggregation, i.e.,
    $\messagee_{\neigh{u}} = \sum_{v \in \neigh{u}} \alpha_{u, v} \embed_v$,
    where the attention weight $\alpha_{u, v}$ is defined as follows:
    \begin{align*}
        \alpha_{u, v} = \frac{exp(a^T[W\embed_u || W\embed_v])}{\sum_{v'\in \neigh{u}} exp(a^T[W\embed_u || W\embed_v'])}
    \end{align*}
    Here, $a$ is a learnable attention vector, $W$ is a learnable matrix, and $||$ denotes the concatenation operation.
\end{itemize}

\mypara{Updating Operation}
We introduce three popular updating operations.

\begin{itemize}[leftmargin=*]
    \item \mypara{Linear Combination~\cite{SGTHM09}}
    The most basic updating method is to calculate the linear combinations, i.e.,
    \begin{align*}
        \upd_{linear} = \sigma(W_{self}\embed_u + W_{neigh}\messagee_{\neigh{u}})
    \end{align*}
    where $W_{self}$ and $W_{neigh}$ are learned during the training process and $\sigma$ is a non-linear activation function.
    The main issue of the basic method is over-smoothing.
    That is the embeddings of all nodes would be similar to each other after several steps of aggregation.

    \item \mypara{Concatenation~\cite{HYL17}}
    One practical approach to handle the over-smoothing issue is to concatenate the result of the linear combination method with the current node embedding, i.e., $\upd_{concat} = \upd_{linear} || \embed_u$.

    \item \mypara{Interpolation~\cite{PTPV17}}
    Another method is to use the weighted average of the linear combination method and the current embedding for updating, i.e., $\upd_{inter} = \alpha_1 \circ \upd_{linear} + \alpha_2 \circ \embed_u$.
\end{itemize}

\mypara{Implementation of GNN Models}
Typically, each step of message passing is referred to as a \textit{GNN module}, and a GNN model can be implemented by stacking multiple layers of the GNN module and one layer of the softmax module for node classification.
We denote a GNN model by \model, which can take as input the feature matrix \feat and the adjacency matrix \adj of a set of nodes \nodeset, and output a posterior matrix \post.
Here each row of \post is the \textit{posterior} of one node $u \in \nodeset$, which is a vector of entries indicating the probability of node $u$ belonging to a certain class.
All values in each row of \post sum up to $1$ by definition.

\section{Visualization of Classical LPA}
\label{app:visualization_lpa}
\begin{figure}[!ht]
    \centering
    \begin{subfigure}{0.7\columnwidth}
    \includegraphics[width=\textwidth]{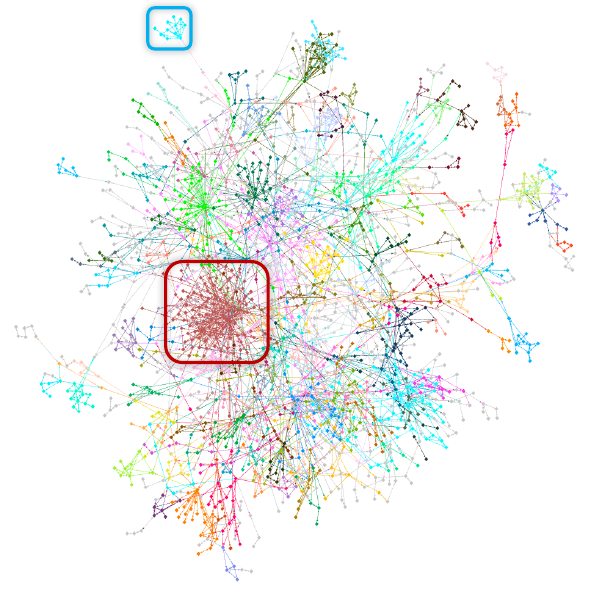}
    \end{subfigure} %
    \caption{
    Visualization of imbalanced shards detected by classical LPA on the Cora dataset.
    Different colors stand for different shards, where the red box and blue box mark a large shard and a small shard, respectively.
    }
    \label{figure:lpa_gephi}
\end{figure}

\autoref{figure:lpa_gephi} shows the visualization of shards detected by classical LPA on the Cora dataset, where different colors stands for different shards.
We use the red box and blue box to mark a large shard and a small shard.
A clear unequal size of the two shards can be observed.

\begin{figure}[!hpbt]
    \centering
    \begin{subfigure}{0.49\columnwidth}
    \includegraphics[width=\textwidth]{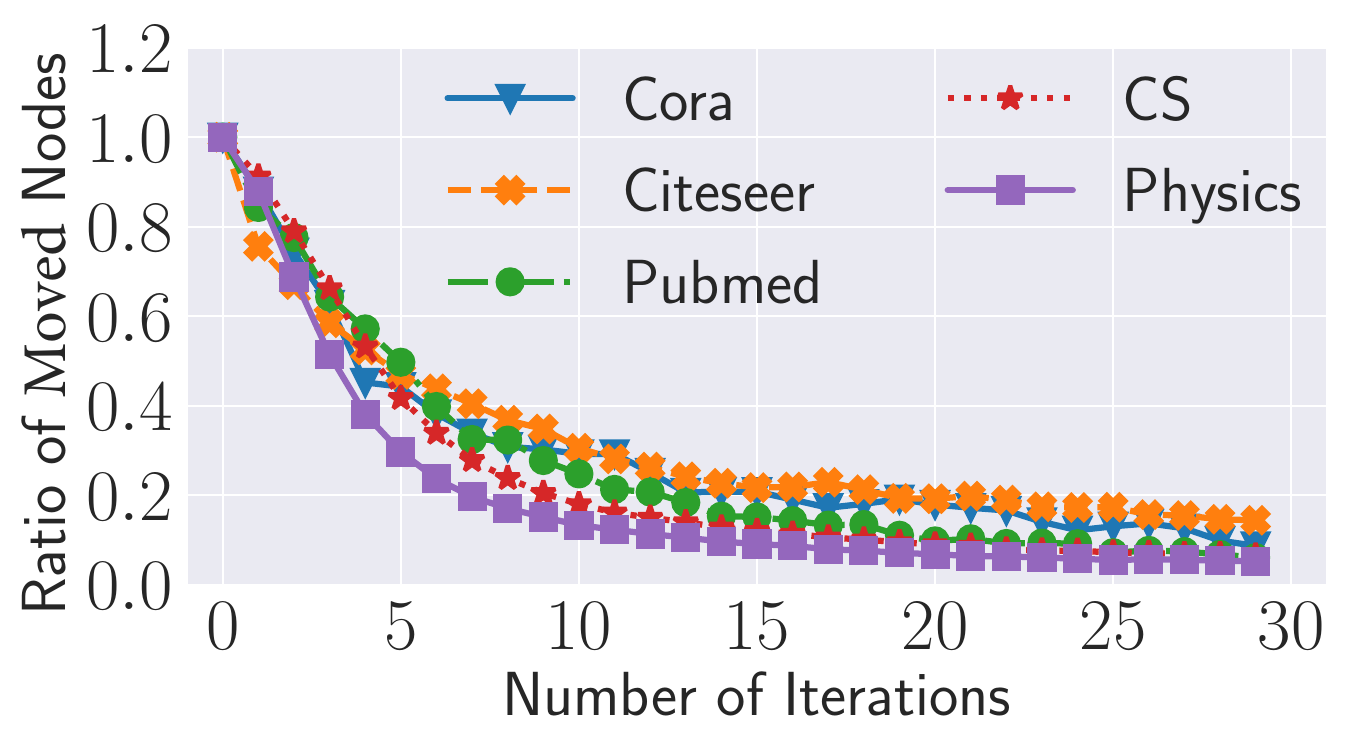} %
    \caption{\blpa}
    \label{subfigure:convergence_blpa}
    \end{subfigure} 
    \begin{subfigure}{0.49\columnwidth}
    \includegraphics[width=\textwidth]{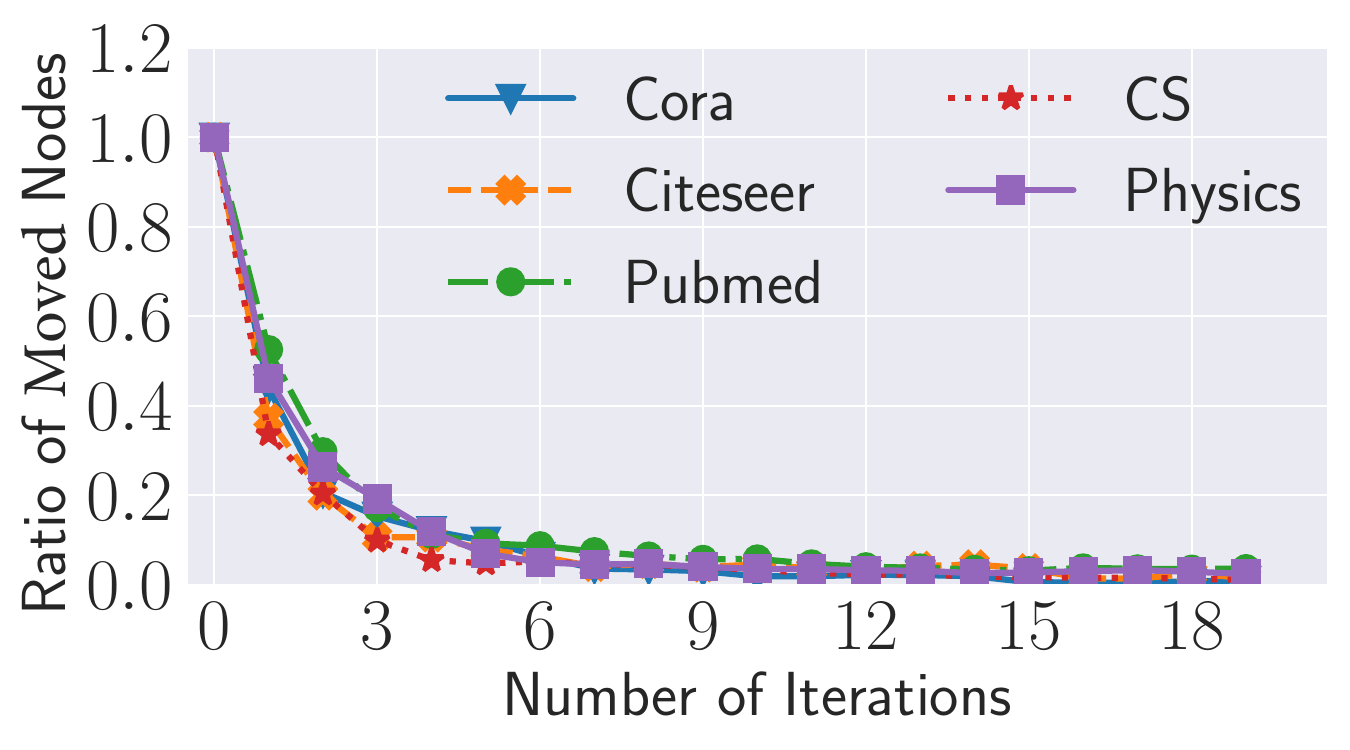} %
    \caption{\bkm}
    \label{subfigure:convergence_bkm}
    \end{subfigure} \\
    \caption{
    Convergence evaluation of \sysname-\blpa and \sysname-\bkm on five datasets.
    }
    \label{figure:convergence}
\end{figure}

\section{Convergence Analysis}
\label{app:convergence}

As discussed in \autoref{section:graph_partition}, it is difficult to theoretically prove the convergence of both \blpa and \bkm.
In this section, we conduct empirical experiments to validate the convergence performance of both algorithms.
An algorithm converges when the nodes of different shards between two consecutive iterations do not move.
\autoref{figure:convergence} illustrates the ratio of moved nodes between different shards in each iteration.
The experimental results show that the ratio of moved nodes gradually approximates to zero within 30 iterations for both algorithms on all five datasets.
Therefore, we set the number of iterations $T$ to 30 for all of our experiments.

\section{Correlation between Importance \\Scores and Shard Properties}
\label{app:shard_importance}

To support the evidence of effectiveness of \optaggr in \autoref{subsection:optaggr_effectiveness}, we next investigate the influence of a shard's properties on its importance score determined by the \optaggr method.

\begin{figure}[!ht]
    \centering
    \begin{subfigure}{0.49\columnwidth}
    \includegraphics[width=\textwidth]{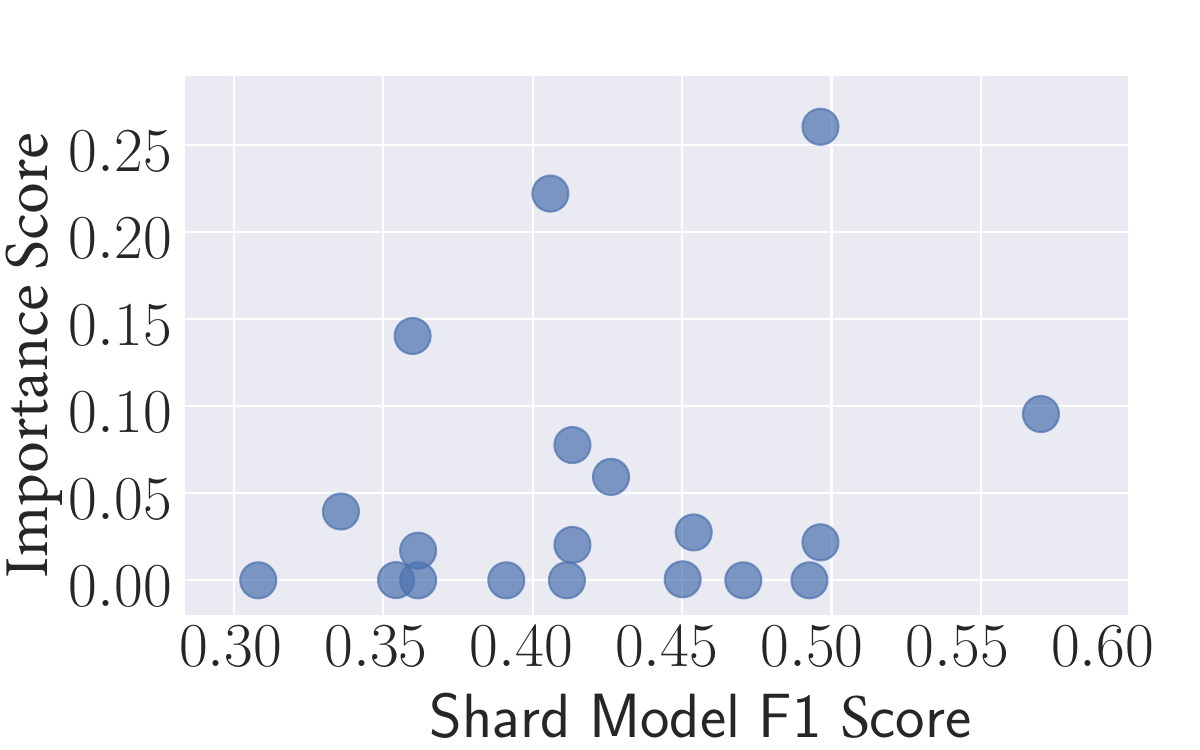}
    \caption{\sysname-\blpa}
    \label{subfigure:blpa_f1_score_vs_weitht}
    \end{subfigure} %
    \begin{subfigure}{0.49\columnwidth}
    \includegraphics[width=\textwidth]{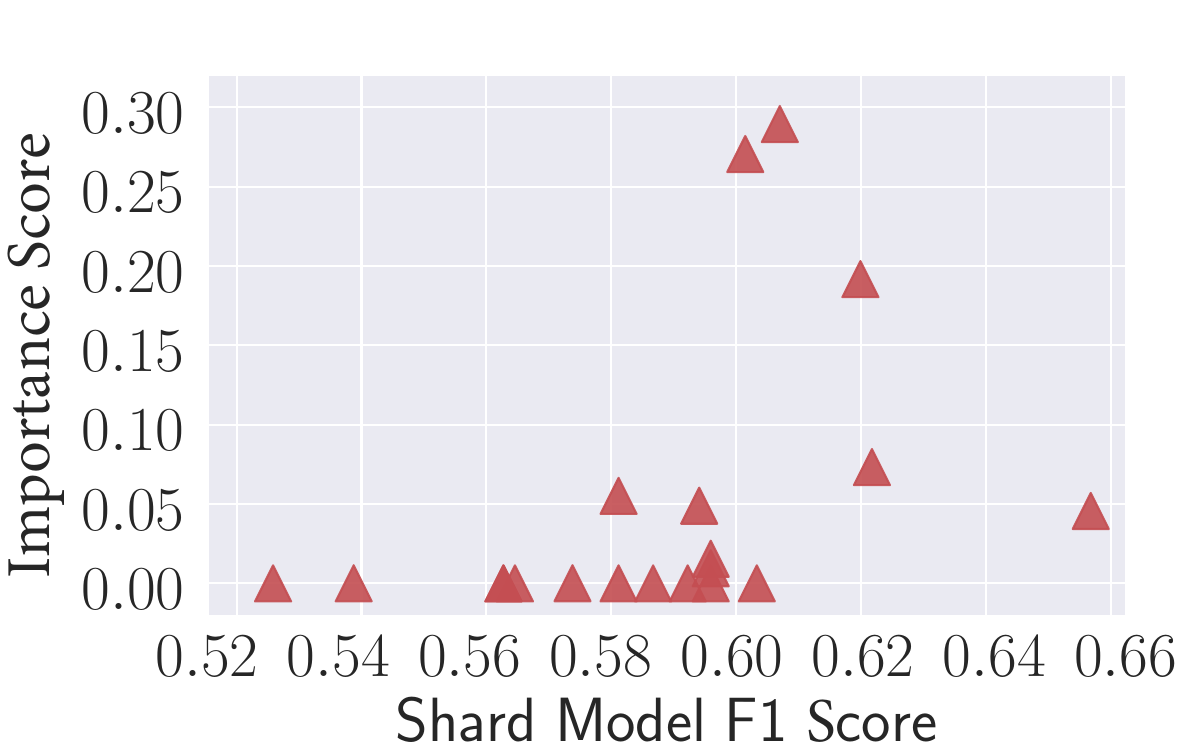}
    \caption{\sysname-\bkm}
    \label{subfigure:bekm_f1_score_vs_weitht}
    \end{subfigure} \\ %
    \caption{
    Correlation between the importance score of a shard model and its F1 score on the Cora dataset.
    The $x$-axis stands for the shard model's F1 score, and the $y$-axis stands for the importance score of that shard. 
    We report the results of GAT model with \sysname-\blpa and \sysname-\bkm unlearning methods.
    }
    \label{figure:f1_score_vs_weitht}
\end{figure}

\autoref{figure:f1_score_vs_weitht} depicts the correlation between each shard's F1 score and its importance score.
Generally, shard models with more accurate predictions are assigned more significant importance scores.
This demonstrates that \optaggr guides \sysname to choose the shards with the highest prediction capability.

\begin{figure}[!ht]
    \centering
    \begin{subfigure}{0.48\columnwidth}
    \includegraphics[width=\textwidth]{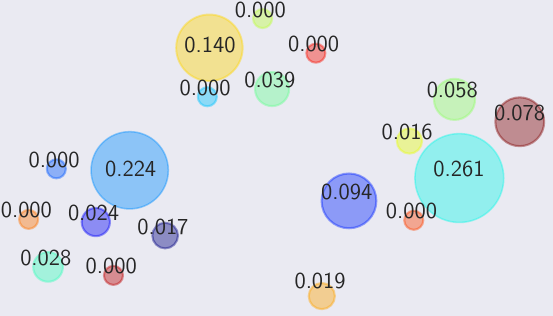}
    \caption{\blpa}
    \label{subfigure:shard_tsne_lpa_GAT}
    \end{subfigure}
    \begin{subfigure}{0.48\columnwidth}
    \includegraphics[width=\textwidth]{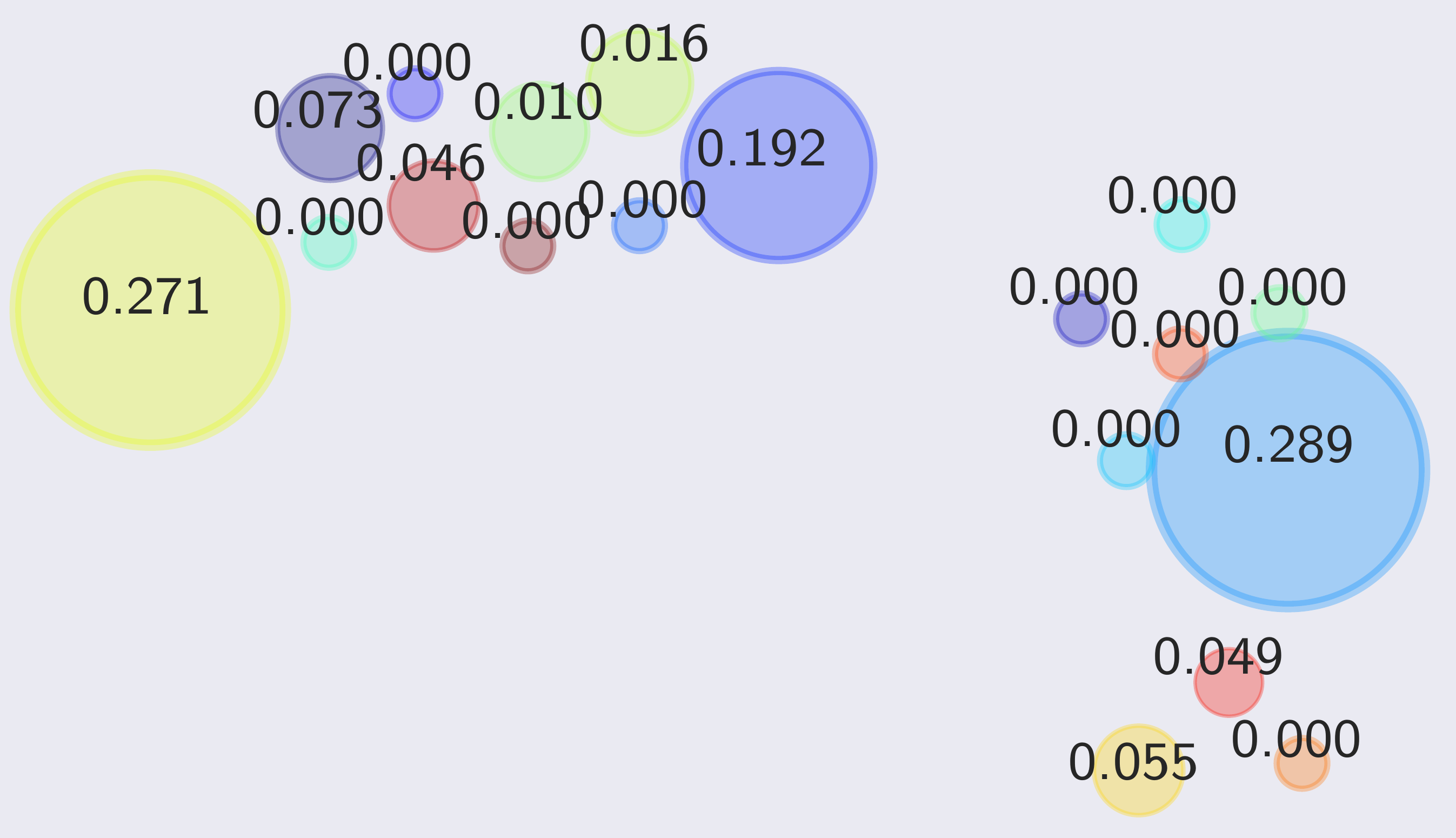}
    \caption{\bkm}
    \label{subfigure:shard_tsne_sagekm_GAT}
    \end{subfigure} \\ %
    \caption{
    The t-SNE plot of shard embeddings for the Cora dataset.
    Each circle represents the mean node embeddings of a shard, where the circle size is proportional to its importance score in annotations. 
    }
    \label{figure:tsne_vs_weitht}
\end{figure}

We further investigate whether each shard's graph properties influence its importance score.
To this end, we extract each shard's embedding by averaging all its nodes' embeddings obtained from the pretrained GNN model and project the shard embedding into a two-dimensional space using t-distributed stochastic neighbor embedding (t-SNE)~\cite{MH08}.
The results are plotted in \autoref{figure:tsne_vs_weitht}.
As we can see, shards with more significant importance scores are typically accompanied by shards with more miniature importance scores.
This implies that for shards trained on similar graphs (similar shard embeddings in the two-dimensional space), our learning-based aggregation assigns a higher score to one of them.
In another way, it also learns to discard redundant information to improve utility.

\section{Role of Graph Structure}
\label{app:role_graph_structure}
\begin{figure*}[!ht]
    \centering
    \begin{subfigure}{2\columnwidth}
    \includegraphics[width=\textwidth]{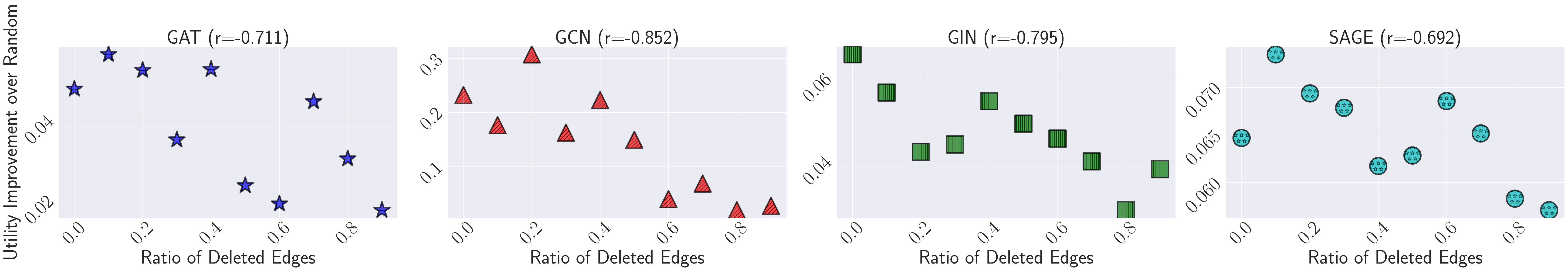} 
    \\ [-3ex]
    \caption{Cora}
    \label{subfigure:cora_gap}
    \end{subfigure} \\ %
    \begin{subfigure}{2\columnwidth}
    \includegraphics[width=\textwidth]{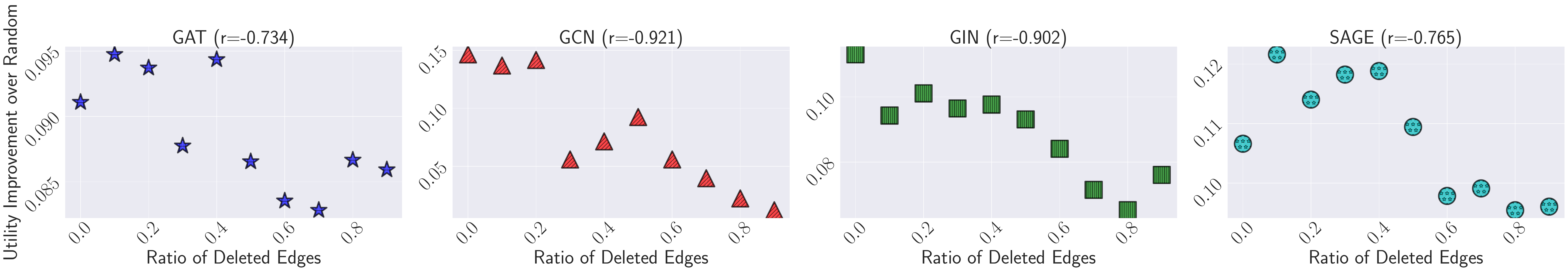} \\ %
    \caption{Citeseer}
    \label{subfigure:citeseer_gap}
    \end{subfigure}
    \caption{
    Correlation between the importance of the graph structure (larger ratio of edge deletion indicates graph structure is less important) and the utility improvement of \sysname over \random.
    }
    \label{figure:gap}
\end{figure*}

To better illustrate the correlation between the importance of the graph structure and the utility improvement of \sysname over \random (\sisa), we performed another experiment on Cora and Citeseer.
Concretely, we delete different fractions of edges from the training graph to model the significance of the graph structure, and then compare the performance gap between \sysname and \random.
\autoref{figure:gap} illustrates the experimental results.
We vary the ratio of deleted edges (as shown in the $x$-axis) from $0\%$ to $90\%$ with a step of 10\%.  
A higher deletion ratio reduces more graph structure information.  
And the $y$-axis stands for the utility improvement of \sysname over \random.  
Although there are some outliers, the overall trend (measured by the Pearson correlation score above each sub-figure) shows that when the graph structure is more important, the utility improvement of \sysname over \random is more significant in most of the cases.

\section{Robustness of \sysname}
\label{app:unlearning_robustness}
In this section, we investigate the impact of the number of unlearned nodes on the model utility of \sysname.
We consider two distributions of node unlearning request: Uniform and non-uniform.
For the uniform unlearning, we randomly delete nodes from all the shards.
For non-uniform, we only delete nodes from half the shards with larger sizes.

\begin{figure*}[!t]
    \centering
    \begin{subfigure}{2\columnwidth}
    \includegraphics[width=\textwidth]{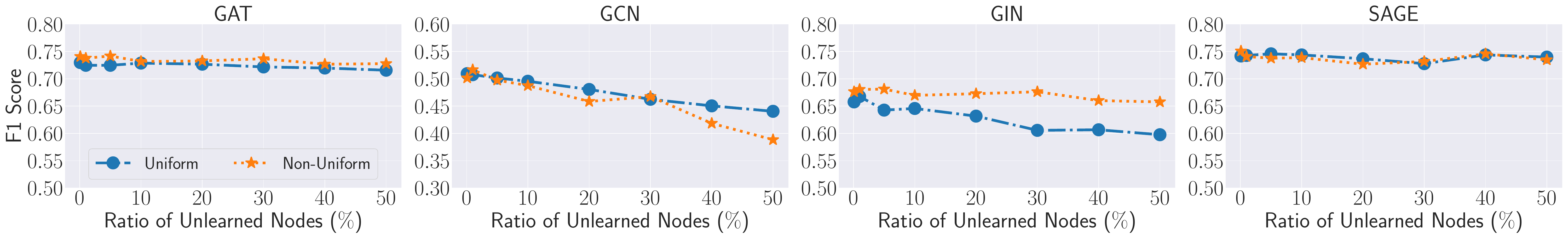} %
    \caption{Citeseer}
    \label{subfigure:ratio_citeseer}
    \end{subfigure}  \\ %
    \begin{subfigure}{2\columnwidth}
    \includegraphics[width=\textwidth]{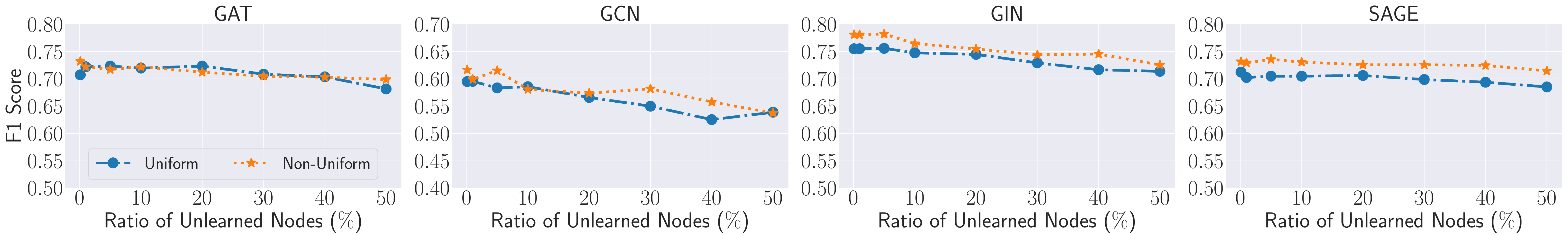} %
    \caption{Cora}
    \label{subfigure:ratio_cora}
    \end{subfigure}  \\ %
    \begin{subfigure}{2\columnwidth}
    \includegraphics[width=\textwidth]{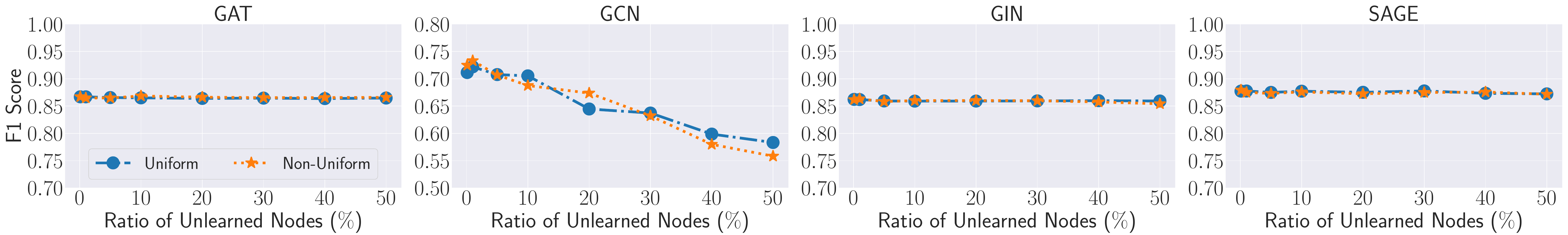} %
    \caption{Pubmed}
    \label{subfigure:ratio_pubmed}
    \end{subfigure} \\
    \caption{
    Impact of the ratio of unlearned nodes on the model utility. We evaluate on both uniform and non-uniform unlearning requests distribution.
    }
    \label{figure:unlearning_adaptive}
\end{figure*}

\autoref{figure:unlearning_adaptive} illustrates the experimental results on three datasets.
We first observe that the F1 scores of \sysname do not drop significantly in most of the settings when the ratio of unlearned nodes is less than 10\%.
When a larger ratio of nodes are deleted, we do observe utility degradation in certain cases.
For instance, for GCN trained on Pubmed, when the ratio of deleted nodes are 50\%, the utility drops from $0.72$ to $0.56$.
Note that in practice, it is unlikely to happen that 50\% of the nodes are deleted.
In general, we conclude that \sysname is robust to a large number of nodes' deletion.
Comparing the results of non-uniform and uniform unlearning, we further observe that the distributions of the deletion do not significantly affect the robustness.

\newpage

\section{Ablation Study}
\label{app:ablation}
We now evaluate the impact of hyperparameters in the graph partitioning phase with regard to the performance of \sysname.

\mypara{Number of Shards $k$}
We conduct the experiments on the Physics dataset with four GNN models.
We vary the number of shards from 2 to 100.    
As suggested in \autoref{subsection:model_utility_evaluation}, we apply \sysname-\bkm for GIN, GAT and SAGE, and \sysname-\blpa for GCN.

\begin{figure}[!ht]
    \centering
    \begin{subfigure}{0.49\columnwidth}
    \includegraphics[width=\textwidth]{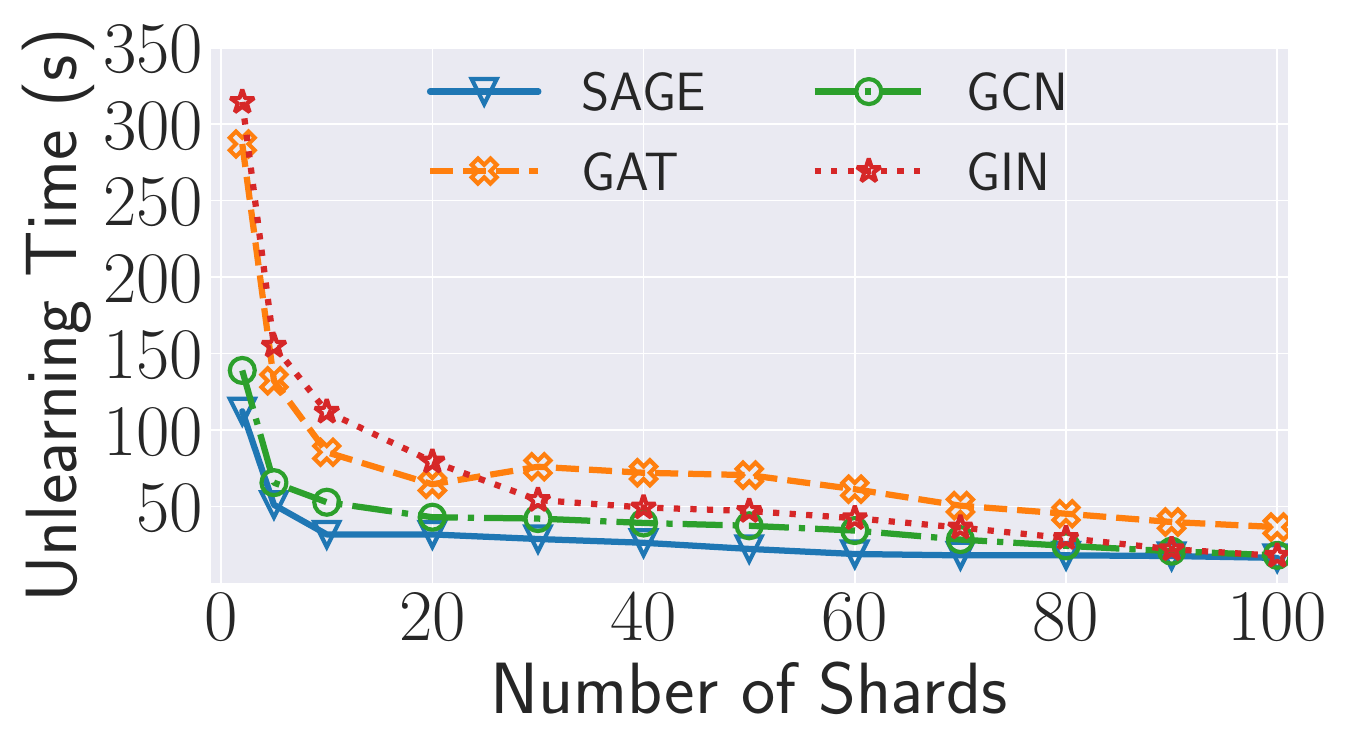} 
    \caption{Efficiency}
    \label{subfigure:num_of_shards_efficiency}
    \end{subfigure} %
    \begin{subfigure}{0.49\columnwidth}
    \includegraphics[width=\textwidth]{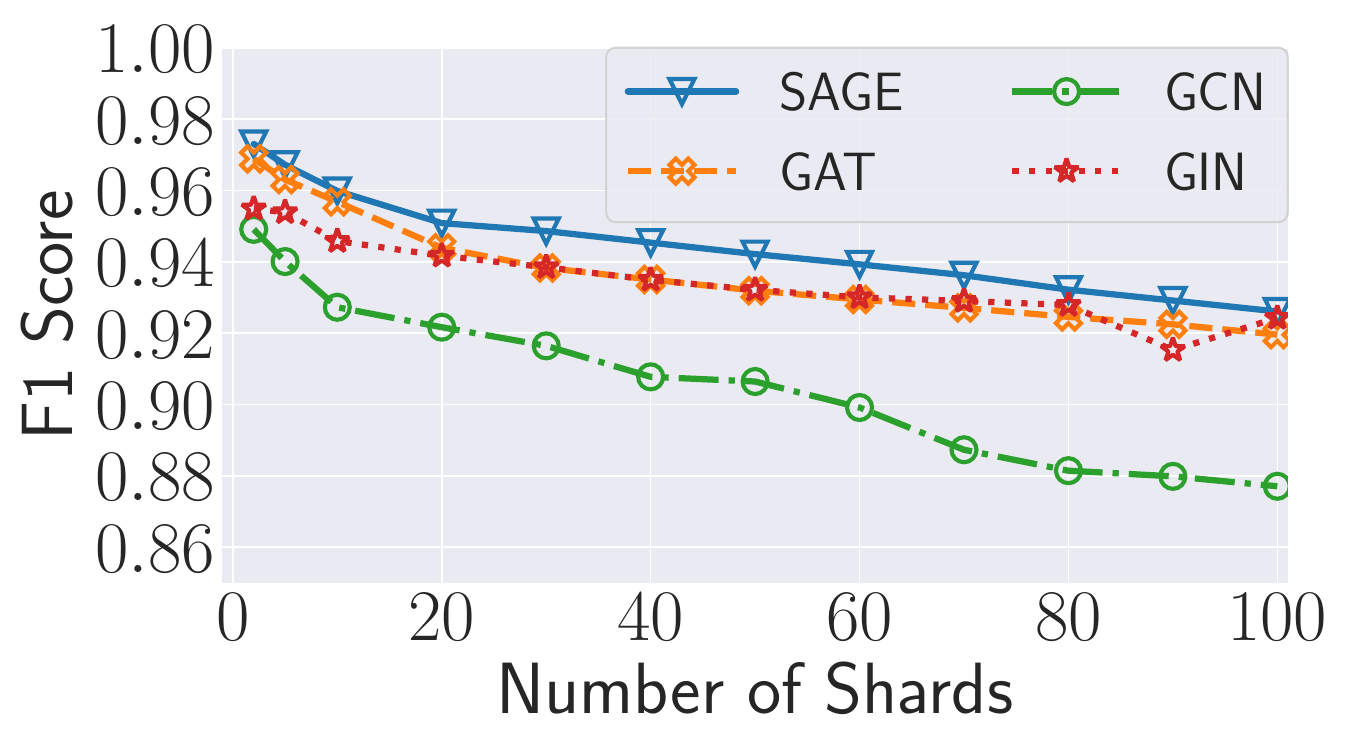} 
    \caption{Utility}
    \label{subfigure:num_of_shards_utility}
    \end{subfigure} \\ %
    \caption{
    Impact of the number of shards $k$ on the unlearning efficiency and model utility on the Physics dataset.
    }
    \label{figure:num_of_shards}
\end{figure}

The results in \autoref{figure:num_of_shards} show that the average unlearning time cost decreases when the number of shards increases for all the GNN models.
This is expected since larger number of shards means smaller shard size, leading to higher unlearning efficiency.
On the other hand, the F1 score of all the four GNN models slightly decrease.
Comparing the four GNN models, the utility of \gcn model drops the most.
We suspect this is because the \gcn model requires the node degree information for normalization which is severely reduced by the graph partitioning.
The number of shards is an important hyperparameter for \sysname, in practice, it should be selected based on the size of the training graph.

\begin{figure}[!ht]
    \centering
    \begin{subfigure}{0.49\columnwidth}
    \includegraphics[width=\textwidth]{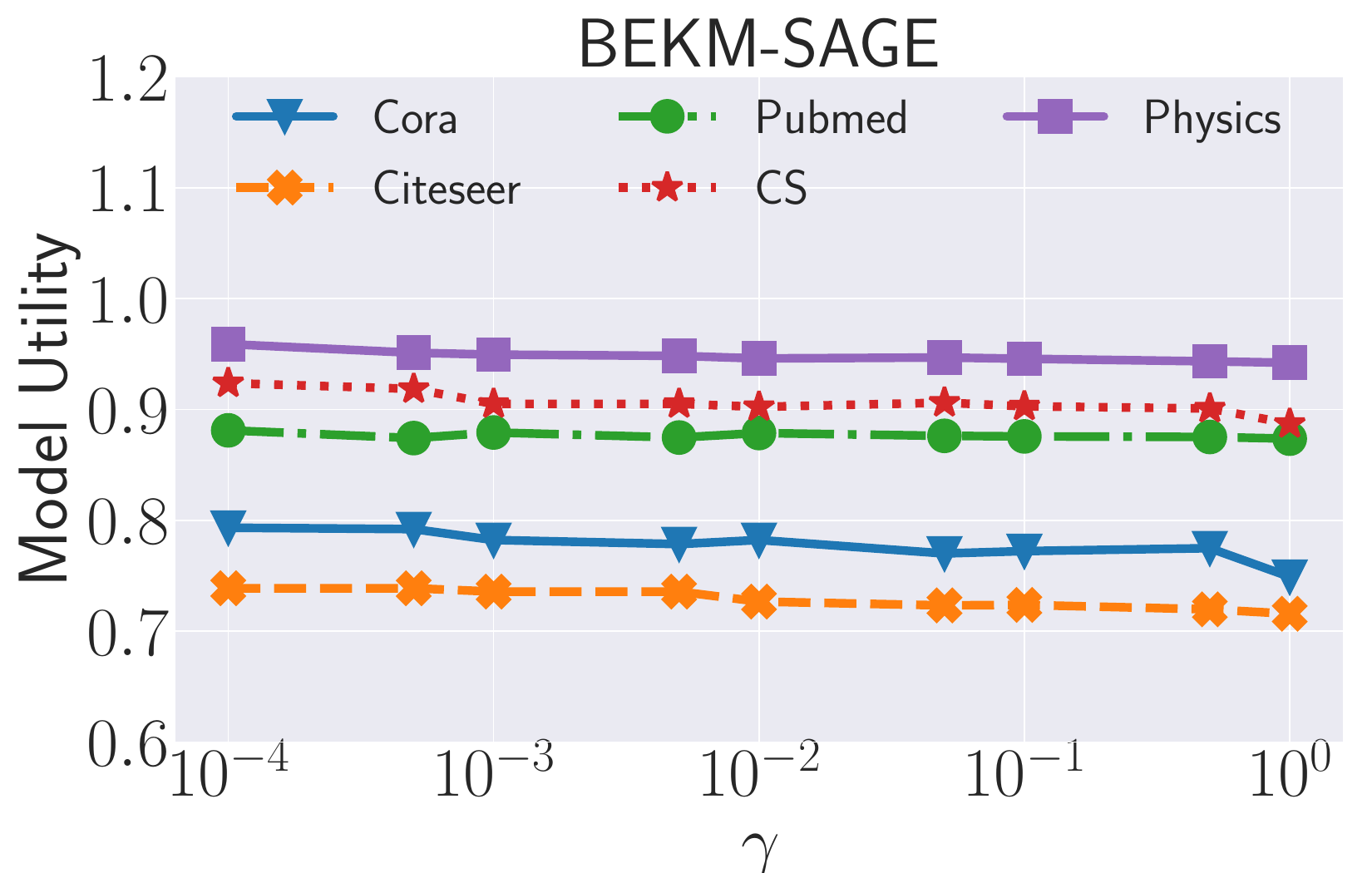} \\ [-2ex]
    \caption{\bkm}
    \label{subfigure:delta_bkm}
    \end{subfigure}
    \begin{subfigure}{0.49\columnwidth}
    \includegraphics[width=\textwidth]{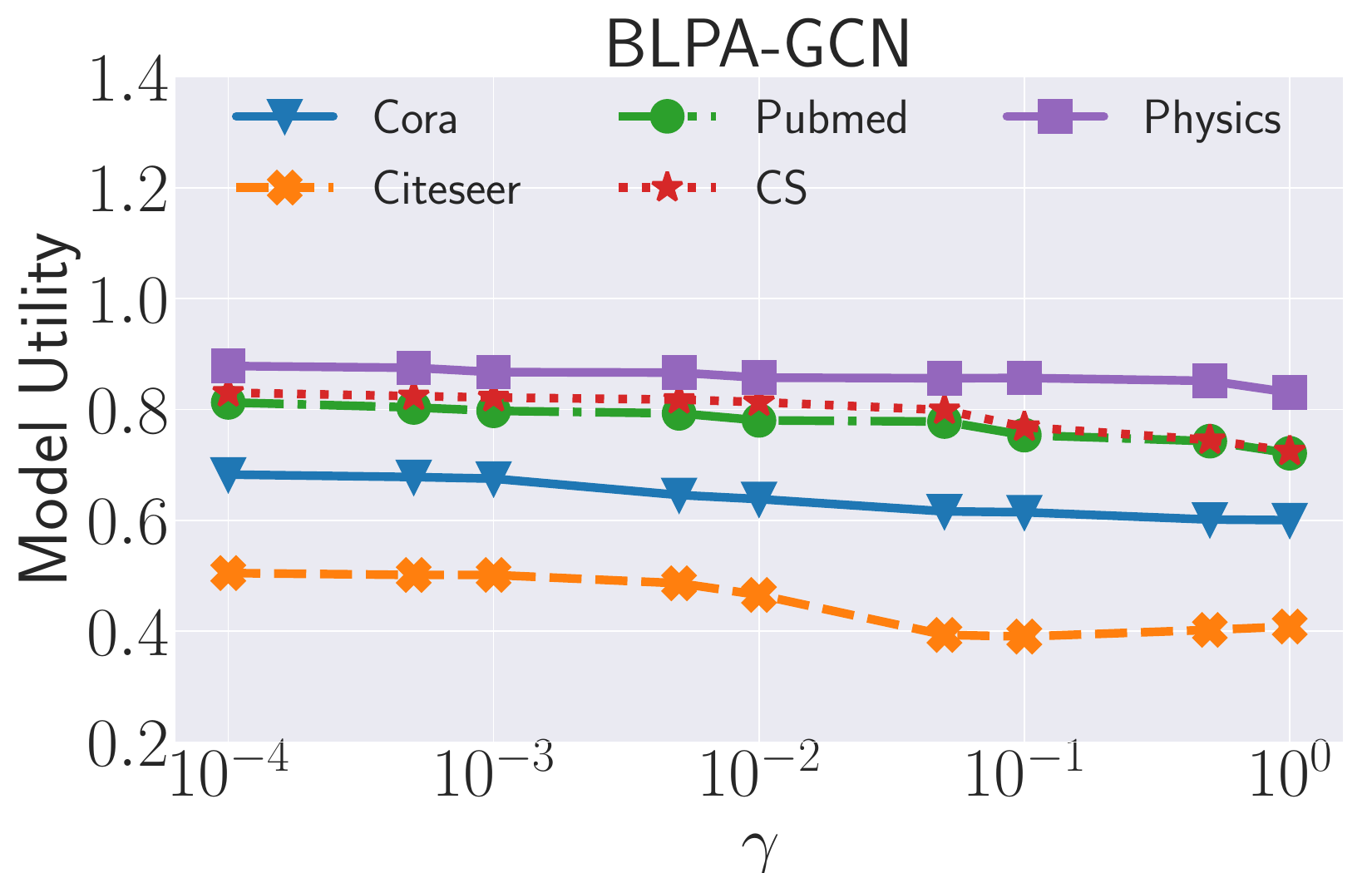} \\ [-2ex]
    \caption{\blpa}
    \label{subfigure:delta_blpa}
    \end{subfigure} \\ %
    \caption{
    Impact of $\delta$ on \sysname-\bkm and \sysname-\blpa for five datasets.
    }
    \label{figure:delta}
\end{figure}

\mypara{Maximum Number of Nodes in Each Shard $\delta$}
It is an important parameter in both \sysname-\blpa and \sysname-\bkm for controlling the degree of balance of the partitioned graphs.
The minimum value of $\delta$ is $\ceil{\frac{n}{k}}$, in which case the shards are balanced.
The maximum value of $\delta$ is $n$, meaning no constraints are enforced to the shard size.
In these cases, \sysname-\blpa and \sysname-\bkm fall back to the standard \lpa and \ekm (embedding clustering with original k-means), respectively.

Intuitively, we aim to make $\delta$ as close as $\ceil{\frac{n}{k}}$ to achieve balanced shards for efficiency.
The remaining concern is what is the impact of $\delta$ on the model utility.
To this end, we conduct experiments for both \sysname-\blpa and \sysname-\bkm on five datasets.
To make the experiments across different datasets comparable, we introduce a scaling parameter $\gamma$ in the range of $[0, 1]$ to regulate the choice of $\delta$, i.e., $\delta = \ceil{\frac{n}{k}} + \gamma \cdot \left( n - \ceil{\frac{n}{k}} \right)$.
When $\gamma=0$, $\delta$ equals to $\ceil{\frac{n}{k}}$, which is the lower bound of $\delta$; when $\gamma=1$, $\delta$ equals to $n$, which is the upper bound of $\delta$.
\autoref{figure:delta} illustrates the experimental results.
In general, we observe that $\delta$ only has a slight impact on the model utility; thus, we set $\delta = \ceil{\frac{n}{k}}$ for all of our experiments which leads to the best efficiency.

\section{Additional Experimental Results on Edge Unlearning}
\label{app:edge_unlearning_results}
\begin{figure*}
    \begin{center}
    \includegraphics[width=0.9\textwidth]{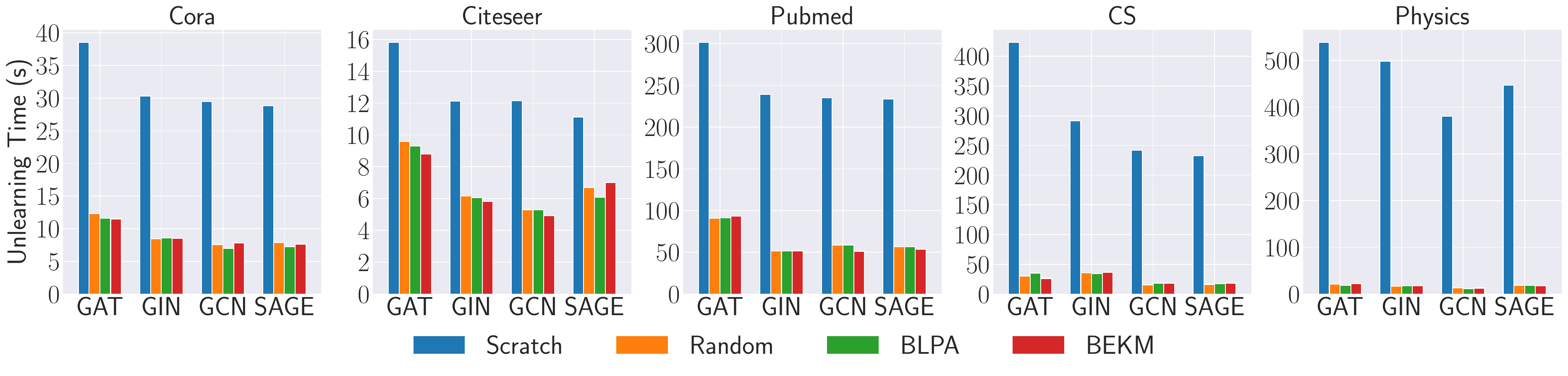}
    \end{center}
    \caption{
    Comparison of edge unlearning efficiency for all graph unlearning methods.
    \blpa and \bkm stand for \sysname-\blpa and \sysname-\bkm unlearning methods, respectively. We observe a similar trend as node unlearning, see \autoref{figure:node_unlearning_efficiency}.
    } 
    \label{figure:edge_unlearning_efficiency}
\end{figure*}

Similar to node unlearning, we conduct experiments to evaluate the unlearning efficiency (corresponding to \autoref{subsection:unlearning_efficiency_evaluation}) and model utility (corresponding to \autoref{subsection:model_utility_evaluation}) for edge unlearning.
\autoref{figure:edge_unlearning_efficiency} and \autoref{table:edge_unlearning_utility_optimal} illustrate the unlearning efficiency and model utility for edge unlearning, respectively.
We reach similar conclusions as node unlearning.

\begin{table}[!ht]
    \centering
    \caption{Comparison of F1 scores of all graph unlearning methods for edge unlearning.
    \blpa and \bkm stand for \sysname-\blpa and \sysname-\bkm unlearning methods, respectively. 
    We highlight our recommended methods in the \colorbox{red!15}{red} ground and the best results in \textcolor{blue}{blue} bold.
    In general, we reach similar conclusions as  \autoref{table:node_unlearning_utility}.}
    \label{table:edge_unlearning_utility_optimal}
    \footnotesize
    \resizebox{\linewidth}{!}{
    \setlength{\tabcolsep}{0.3em}
    \renewcommand{\arraystretch}{1.0}
    \begin{tabular}{c c | c | c c c }
    \toprule
    \textbf{Dataset} & \textbf{Model} & \scratch & \random & \blpa & \bkm \\
    \toprule
    & \textbf{GAT} & 0.823 $\pm$ 0.005 & 0.723 $\pm$ 0.009 & \textbf{\textcolor{blue}{0.774 $\pm$ 0.008}} & \cellcolor{red!15}0.756 $\pm$ 0.005 \\
    & \textbf{GCN} & 0.742 $\pm$ 0.004 & 0.448 $\pm$ 0.005 & \cellcolor{red!15}\textbf{\textcolor{blue}{0.657 $\pm$ 0.005}} & 0.474 $\pm$ 0.002 \\
    & \textbf{GIN} & 0.786 $\pm$ 0.011 & 0.755 $\pm$ 0.007 & 0.762 $\pm$ 0.009 & \cellcolor{red!15}\textbf{\textcolor{blue}{0.768 $\pm$ 0.027}} \\
    \multirow{-4}{*}{\rotatebox[origin=c]{90}{\textbf{Cora}}}
    & \textbf{SAGE} & 0.827 $\pm$ 0.007 & 0.669 $\pm$ 0.005 & 0.721 $\pm$ 0.003 & \cellcolor{red!15}\textbf{\textcolor{blue}{0.731 $\pm$ 0.002}} \\
    \midrule
    & \textbf{GAT} & 0.706 $\pm$ 0.003 & 0.620 $\pm$ 0.017 & \textbf{\textcolor{blue}{0.674 $\pm$ 0.002}} & \cellcolor{red!15}0.670 $\pm$ 0.001 \\
    & \textbf{GCN} & 0.470 $\pm$ 0.005 & 0.464 $\pm$ 0.004 & \cellcolor{red!15}0.532 $\pm$ 0.008 & \textbf{\textcolor{blue}{0.571 $\pm$ 0.017}} \\
    & \textbf{GIN} & 0.610 $\pm$ 0.019 & 0.592 $\pm$ 0.015 & 0.632 $\pm$ 0.026 & \cellcolor{red!15}\textbf{\textcolor{blue}{0.736 $\pm$ 0.020}} \\
    \multirow{-4}{*}{\rotatebox[origin=c]{90}{\textbf{Citeseer}}}
    & \textbf{SAGE} & 0.667 $\pm$ 0.002 & 0.670 $\pm$ 0.012 & 0.680 $\pm$ 0.062 & \cellcolor{red!15}\textbf{\textcolor{blue}{0.711 $\pm$ 0.006}} \\
    \midrule
    & \textbf{GAT} & 0.844 $\pm$ 0.003 & 0.827 $\pm$ 0.002 & 0.848 $\pm$ 0.002 & \cellcolor{red!15}\textbf{\textcolor{blue}{0.854 $\pm$ 0.007}} \\
    & GCN & 0.740 $\pm$ 0.001 & 0.549 $\pm$ 0.005 & \cellcolor{red!15}\textbf{\textcolor{blue}{0.716 $\pm$ 0.010}} & 0.578 $\pm$ 0.002 \\
    & \textbf{GIN} & 0.846 $\pm$ 0.015 & 0.857 $\pm$ 0.050 & 0.865 $\pm$ 0.004 & \cellcolor{red!15}\textbf{\textcolor{blue}{0.859 $\pm$ 0.003}} \\
    \multirow{-4}{*}{\rotatebox[origin=c]{90}{\textbf{Pubmed}}}
    & \textbf{SAGE} & 0.873 $\pm$ 0.001 & 0.837 $\pm$ 0.002 & \textbf{\textcolor{blue}{0.868 $\pm$ 0.002}} & \cellcolor{red!15}0.855 $\pm$ 0.002 \\
    \midrule
    & \textbf{GAT} & 0.930 $\pm$ 0.004 & 0.882 $\pm$ 0.010 & 0.847 $\pm$ 0.002 & \cellcolor{red!15}\textbf{\textcolor{blue}{0.896 $\pm$ 0.001}} \\
    & \textbf{GCN} & 0.905 $\pm$ 0.006 & 0.706 $\pm$ 0.018 & \cellcolor{red!15}\textbf{\textcolor{blue}{0.790 $\pm$ 0.003}} & 0.732 $\pm$ 0.022 \\
    & \textbf{GIN} & 0.887 $\pm$ 0.005 & 0.858 $\pm$ 0.005 & 0.789 $\pm$ 0.013 & \cellcolor{red!15}\textbf{\textcolor{blue}{0.861 $\pm$ 0.002}} \\
    \multirow{-4}{*}{\rotatebox[origin=c]{90}{\textbf{CS}}}
    & \textbf{SAGE} & 0.953 $\pm$ 0.004 & 0.898 $\pm$ 0.009 & 0.896 $\pm$ 0.015 & \cellcolor{red!15}\textbf{\textcolor{blue}{0.923 $\pm$ 0.001}} \\
    \midrule
    & \textbf{GAT} & 0.956 $\pm$ 0.002 & 0.910 $\pm$ 0.003 & 0.925 $\pm$ 0.002 & \cellcolor{red!15}\textbf{\textcolor{blue}{0.928 $\pm$ 0.003}} \\
    & \textbf{GCN} & 0.942 $\pm$ 0.005 & 0.729 $\pm$ 0.013 & \cellcolor{red!15}\textbf{\textcolor{blue}{0.853 $\pm$ 0.007}} & 0.773 $\pm$ 0.002 \\
    & \textbf{GIN} & 0.939 $\pm$ 0.003 & 0.910 $\pm$ 0.005 & 0.917 $\pm$ 0.003 & \cellcolor{red!15}\textbf{\textcolor{blue}{0.929 $\pm$ 0.002}} \\
    \multirow{-4}{*}{\rotatebox[origin=c]{90}{\textbf{Physics}}}
    & \textbf{SAGE} & 0.950 $\pm$ 0.005 & 0.817 $\pm$ 0.021 & 0.924 $\pm$ 0.001 & \cellcolor{red!15}\textbf{\textcolor{blue}{0.936 $\pm$ 0.001}} \\
    \bottomrule
    \end{tabular}
    }
\end{table}

\section{Community-Dependent Removal \\Requests}
\label{app:community_depedent}
\sysname can deal with the scenario where unlearning requests come from specific types of communities, since it can be seen as the shard-size-constraint version of community detection algorithms, and nodes in specific types of communities tend to be allocated into the same shards. 
When the community-dependent requests come, only a few shard models need to be retrained.

To further validate the robustness of \sysname, we use the vanilla LPA method (without community size constraints) to partition the graph.
We then randomly choose a community detected by vanilla LPA, and gradually delete the corresponding nodes (with the same IDs in the selected community) from the shard models of \sysname.

\begin{figure*}[!t]
    \centering
    \begin{subfigure}{1.9\columnwidth}
    \includegraphics[width=\textwidth]{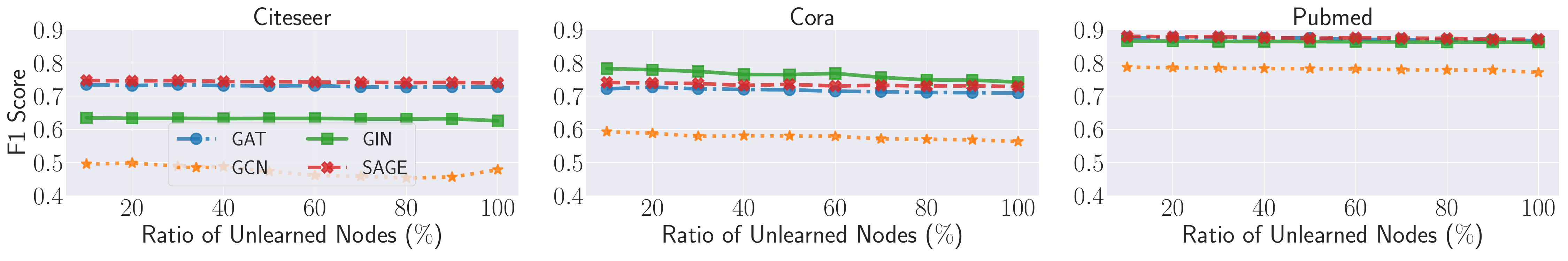}
    \end{subfigure}
    \caption{
    Model utility when the removal requests come from specific community detected by vanilla LPA.
    X-axis stands for the ratio of unlearned nodes in the selected community.}
    \label{figure:dependent_unlearning}
\end{figure*}

\autoref{figure:dependent_unlearning} illustrates the model utility when the removal requests come from specific community detected by vanilla LPA.
The experimental results show that deleting nodes from a single community does not significantly affect the model utility of \sysname.

\end{document}